\PassOptionsToPackage{table}{xcolor}
\documentclass[11pt]{article}
\usepackage{acl}
\usepackage{times}
\usepackage{latexsym}
\usepackage{microtype}
\usepackage{inconsolata}
\usepackage[T1]{fontenc}
\usepackage{amsmath,amssymb,amsthm}
\usepackage{graphicx}
\usepackage{booktabs,tabularx,multirow,colortbl,listings,wrapfig,float,array}
\usepackage[most]{tcolorbox}
\usepackage{xcolor}
\usepackage{tikz}
\usetikzlibrary{arrows.meta,positioning}
\usepackage{url}
\usepackage{enumitem}

\newcommand{\ind}{\mathbf{1}}

\title{RuACT: Recovering Inactive Rubrics as Adaptive Critics and Teachers}

\author{
  {\bfseries
  Yijun Li$^{1}$ \quad
  Zhengxi Lu$^{1}$ \quad
  Zhipiao Liu$^{2}$ \quad
  Jinzheng Yu$^{2}$} \\
  {\bfseries
  Xutao Weng$^{2}$ \quad
  Jian Shao$^{1}$ \quad
  Yongliang Shen$^{1}$} \\
  $^{1}$Zhejiang University
  \hspace{1.5cm}
  $^{2}$Huawei \\
  {\normalfont
   \fontfamily{pcr}\fontseries{m}\fontshape{n}\selectfont
   \{li\_yijun,\hspace{0.9em}zhengxilu,\hspace{0.9em}syl\}@zju.edu.cn
  } \\
}

\usetikzlibrary{arrows.meta,positioning}

\definecolor{promptgray}{RGB}{246,246,246}
\definecolor{placeholderred}{RGB}{180,50,50}

\lstdefinestyle{promptstyle}{
    basicstyle=\ttfamily\small,
    columns=fullflexible,
    keepspaces=true,
    breaklines=true,
    showstringspaces=false,
    moredelim=[is][\color{placeholderred}]{@@}{@@}
}

\newtcblisting{promptbox}{
    enhanced,
    breakable,
    listing only,
    listing engine=listings,
    listing options={style=promptstyle},
    colback=promptgray,
    colframe=black,
    boxrule=0.6pt,
    arc=4mm,
    left=4mm,
    right=4mm,
    top=3.5mm,
    bottom=3.5mm
}

\definecolor{casegray}{RGB}{247,247,247}
\definecolor{caseborder}{RGB}{185,185,185}
\definecolor{casetextgray}{RGB}{90,90,90}

\definecolor{childlight}{RGB}{245,248,250}
\definecolor{childmid}{RGB}{235,241,245}
\definecolor{childdark}{RGB}{223,232,238}

\newtcolorbox{querybox}{
    enhanced,
    colback=casegray,
    colframe=caseborder,
    boxrule=0.55pt,
    arc=2.5mm,
    left=2.8mm,
    right=2.8mm,
    top=2mm,
    bottom=2mm
}

\newtcolorbox{parentbox}{
    enhanced,
    colback=casegray,
    colframe=black!55,
    boxrule=0.7pt,
    arc=2.5mm,
    left=2.8mm,
    right=2.8mm,
    top=2.2mm,
    bottom=2.2mm
}

\newtcolorbox{childbox}[1]{
    enhanced,
    colback=#1,
    colframe=caseborder,
    boxrule=0.55pt,
    arc=2.5mm,
    left=2.4mm,
    right=2.4mm,
    top=2.2mm,
    bottom=2.2mm,
    height=39mm,
    valign=center
}

\definecolor{promptgray}{RGB}{246,246,246}
\definecolor{placeholderred}{RGB}{190,30,30}

\lstdefinestyle{promptcase}{basicstyle=\ttfamily\footnotesize,breaklines=true,breakatwhitespace=true,columns=fullflexible,keepspaces=true,showstringspaces=false,frame=single,framerule=0.3pt,framesep=5pt,xleftmargin=6pt,xrightmargin=6pt,aboveskip=6pt,belowskip=8pt}

\begin{document}
\maketitle

\begin{abstract}
Rubric-based reinforcement learning (RL) extends RL with verifiable rewards to open-ended tasks by scoring each response against query-specific rubrics. Under group-relative optimization, however, a rubric that every sampled response fails (unreachable) or passes (saturated) gives all responses the same outcome and no learning signal; in Qwen3-4B medical training, 60.88\% of query--rubric observations are inactive. We find that the two states lack different supervision: stating an unreachable rubric in the prompt raises its satisfaction from 1.13\% to 91.25\%, whereas responses that all pass a saturated rubric still differ in quality. We therefore propose \textbf{RuACT} (\textbf{Ru}brics as \textbf{A}daptive \textbf{C}ritics and \textbf{T}eachers), which assigns each inactive rubric the role that supplies what it lacks. Rubric as Teacher uses unreachable rubrics as privileged information for reward-gated on-policy self-distillation, and Rubric as Critic decomposes saturated rubrics into child rubrics that separate current responses while sharing the parent's weight. RuACT outperforms GRPO, rubric reweighting, and rubric-guided exploration on all four medical and scientific benchmarks with Qwen3-4B and achieves the best average with Qwen3-1.7B. 
On initially unreachable rubrics, unguided recovery rises from 35.02\% to 42.86\%, while child refinement improves reward–quality alignment on independent rollouts from 57.74\% to 60.77\%.
Code is available at \url{https://github.com/ZJU-REAL/RuACT}.

\end{abstract}

\begin{figure}[t]
  \centering
  \includegraphics[width=\columnwidth]{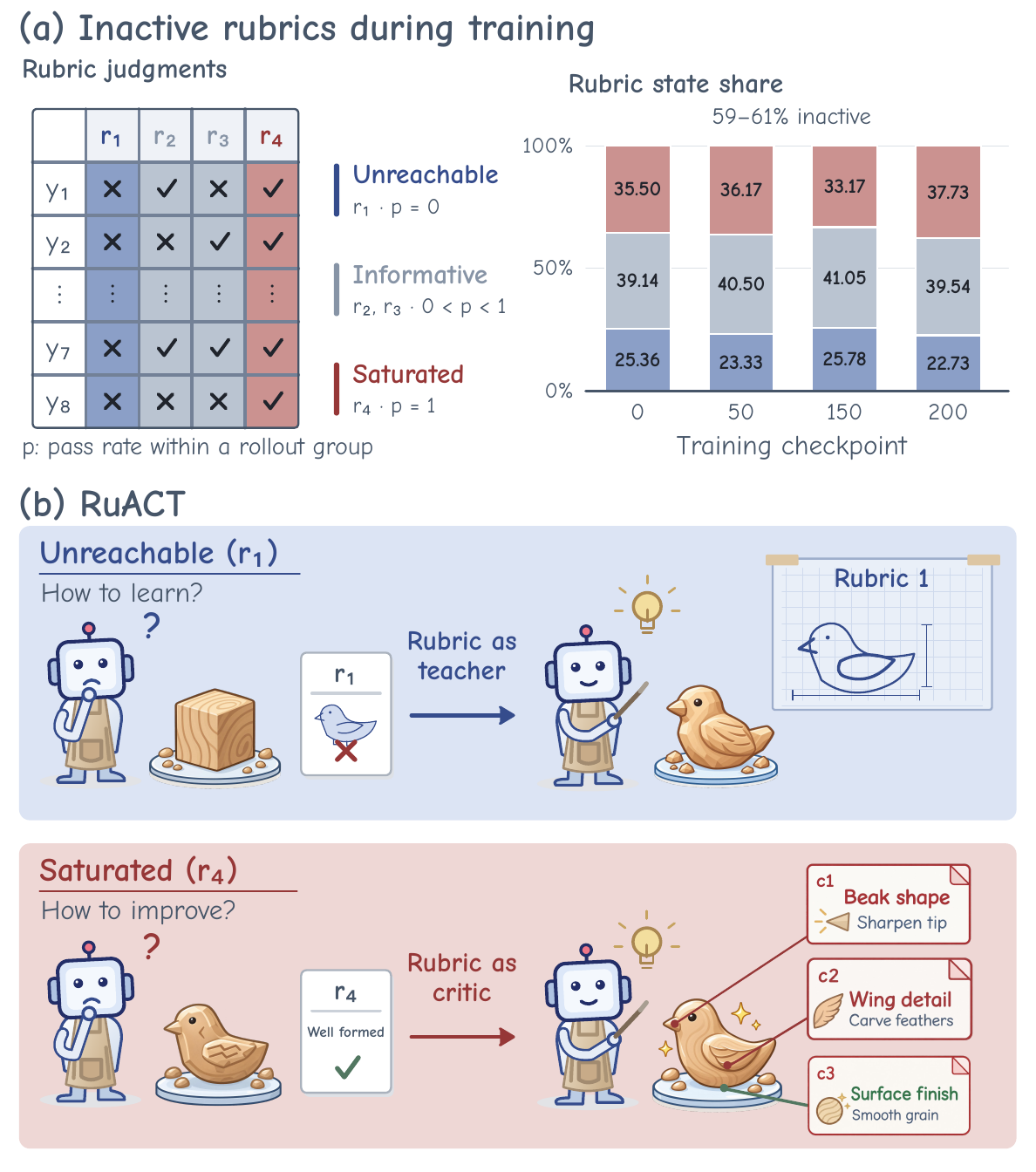}
  \caption{\textbf{Inactive rubrics and RuACT.}
  (a) Illustration of unreachable, informative, and saturated rubrics and their distribution through training. (b) How RuACT recovers inactive rubrics: reward-verified guidance helps elicit missing behaviors, while finer-grained criticism exposes details for further improvement.}
  \label{fig:fig1}
\end{figure}

\section{Introduction}
\label{sec:Introduction}

Rubric-based reinforcement learning (RL) extends the success of RL with verifiable rewards (RLVR) in mathematics and code~\citep{shao2024deepseekmath,guo2025deepseekr1,cheng2026exploration} to open-ended tasks, where response quality is multidimensional and no single verifier suffices, by decomposing the objective into fine-grained rubrics that each yield a reward~\citep{gunjal2026rubrics,huang2025rubricanchors,wei2026qurl,li2026ares,viswanathan2025checklists,he2026advancedif}. Yet a task-relevant rubric is not necessarily useful to the current policy: under group-relative optimization, a rubric that no sampled response satisfies (\emph{unreachable}) or that every response satisfies (\emph{saturated}) assigns the same outcome to all rollouts and cancels out of the advantage~\citep{tyagi2026pow3r}, leaving only rubrics with intermediate pass rates \emph{informative}. We refer to these two endpoint states as \emph{inactive rubrics}. Both are common: in Qwen3-4B medical training, \textbf{25.19\% of query--rubric pairs are unreachable and 35.69\% saturated} (Section~\ref{sec:preliminary}).

Existing methods use rubrics as \emph{critics}, prioritizing informative feedback through reweighting and curricula~\citep{tyagi2026pow3r,chen2026rucl}, or continually updating rubrics to sustain reward contrast~\citep{shao2025drtulu,rezaei2025onlinerubrics,ding2026evorubrics,xu2026sibylsense}.
Yet reweighting and curricula cannot restore within-group contrast for an inactive rubric, because its constant contribution cancels under any weight (Section~\ref{sec:prelim_grpo}); rubric updates may improve discrimination, but they replace the original requirements rather than recover their supervision.
Complementarily, rubrics serve as \emph{teachers} by guiding exploration~\citep{zhou2025ruscarl,zhang2026herl} or providing privileged information for self-distillation~\citep{xia2026cripo,rezaei2026rgsd,gu2026rcsd,ye2026opcd}.
Such guidance can elicit missing behaviors, but teaching existing requirements does not itself sharpen their evaluation.
Once a coarse requirement is satisfied, further improvements may remain indistinguishable under its original judgments.

The two endpoints, however, lack different kinds of supervision (Section~\ref{sec:preliminary}). For an unreachable rubric, the behavior is missing from ordinary rollouts: stating the rubric in the prompt raises the satisfaction rate of a frozen Qwen3-4B from 1.13\% to 91.25\%. For a saturated rubric, the distinction is missing: an independent fine-grained judge finds quality differences among its responses in 71\% of cases, compared with 29\% for unreachable rubrics. Teaching thus supplies behaviors but not contrast, whereas finer evaluation supplies contrast where the behaviors already exist. \emph{Each inactive rubric should therefore take the role that supplies what it lacks: a teacher when no response satisfies it, and a finer-grained critic when every response does.}

Following this principle, we propose \textbf{RuACT} (\textbf{Ru}brics as \textbf{A}daptive \textbf{C}ritics and \textbf{T}eachers), which assigns each rubric a role by its current state in each rollout group; informative rubrics keep their original rewards. \textbf{Rubric as Teacher} appends unreachable rubrics to the prompt of a frozen self-teacher and distills its token distributions on the best ordinary rollout into the unguided policy. To avoid trading off other rubrics, distillation is enabled only when a rubric-guided response outscores every ordinary rollout. \textbf{Rubric as Critic} decomposes each saturated rubric into child rubrics, retains those that separate current rollouts, and lets them share the parent's reward weight, which restores contrast while preserving the original objective.

On four medical and scientific benchmarks, RuACT outperforms GRPO, rubric reweighting, and rubric-guided exploration on every benchmark with Qwen3-4B, with the largest gain on ResearchQA (from 64.82\% under GRPO to 69.44\%), and achieves the best average with Qwen3-1.7B. Without guidance, the trained policy satisfies 42.86\% of initially unreachable rubrics (35.02\% under GRPO), and refined rewards agree better with an independent judge (from 57.74\% to 60.77\%). The gains persist under a second judge (Appendix~\ref{app:judge-robustness}).

Our contributions are threefold:
\begin{itemize}[itemsep=1pt,topsep=2pt,parsep=0pt]
    \item We show that inactive rubrics cover 60.88\% of query--rubric observations, and that unreachable rubrics lack behaviors while saturated rubrics lack distinctions among responses.

    \item We propose \textbf{RuACT}, which assigns each inactive rubric the role that supplies its missing supervision: a teacher for unreachable rubrics and a critic for saturated ones.

    \item Across two domains and two model scales, RuACT improves over strong rubric-based baselines, and analyses confirm that each role recovers the supervision it targets.
\end{itemize}

\section{Preliminary}
\label{sec:preliminary}

\subsection{Rubric-Based GRPO}
\label{sec:prelim_grpo}

Given query $x$, GRPO~\citep{shao2024deepseekmath} samples $G$ responses
$\{y_i\}_{i=1}^{G}$ from $\pi_{\theta_{\rm old}}(\cdot\mid x)$ and
evaluates them against $\mathcal R_x=\{(r_j,w_j)\}_{j=1}^{J_x}$.
Let $e_{ij}\in\{0,1\}$ denote the judgment of response $y_i$ under rubric $r_j$. With $Z_x=\sum_{j:w_j>0}w_j$, the reward and advantage are
\begin{equation}
R_i=\frac{\sum_j w_j e_{ij}}{Z_x},
\qquad
A_i=\frac{R_i-\overline R}{\sigma(R)+\epsilon},
\label{eq:reward}
\end{equation}
where $\overline R$ and $\sigma(R)$ are the group reward mean and standard deviation.
When rubric $r_j$ receives identical judgments across all $G$ responses, its contribution is constant and cancels during reward centering; its weight still enters $Z_x$, but this rescales $R_i-\overline R$ and $\sigma(R)$ alike. Reweighting, as in our Reweight baseline, can therefore redistribute informative rubrics but cannot create within-group contrast from an inactive one, up to $\epsilon$.

\subsection{Inactive Rubrics}
\label{sec:prelim_inactive}

For a rubric with nonzero weight, let $s_{ij}=e_{ij}$ if $w_j>0$ and $s_{ij}=1-e_{ij}$ otherwise, so $s_{ij}=1$ means $y_i$ meets its intent. The satisfaction rate is
\begin{equation}
\widehat p_j=\frac{1}{G}\sum_{i=1}^{G}s_{ij}.
\label{eq:rubric_pass_rate}
\end{equation}
As illustrated in Figure~\ref{fig:fig1}(a), rubrics are \emph{unreachable} when $\widehat p_j=0$, \emph{informative} when $0<\widehat p_j<1$, and \emph{saturated} when $\widehat p_j=1$.
The endpoints are \emph{inactive}, since their constant contributions cancel (Section~\ref{sec:prelim_grpo}). Across 122,768 query--rubric observations during Qwen3-4B GRPO training on RaR-Medicine, 25.19\% are unreachable and 35.69\% saturated, totaling 60.88\% inactive observations. The fraction lies between 58.95\% and 60.86\% at each checkpoint shown in Figure~\ref{fig:fig1}(a). Inactive rubrics are thus pervasive in rubric-based GRPO, and the two states co-occur: 75.06\% of query groups contain both unreachable and saturated rubrics (Appendix~\ref{app:inactivity-statistics}), so a single loop must handle both.

\subsection{Rubrics as Teachers and Critics}
\label{sec:prelim_recovery}

We probe the two endpoints with inference-only diagnostics on RaR-Medicine using the frozen Qwen3-4B base model.
For each query, we first sample $G=8$ ordinary responses and classify each query--rubric pair as unreachable, informative, or saturated based on its binary judgments.

\begin{figure}[t]
\centering
\includegraphics[width=\linewidth]{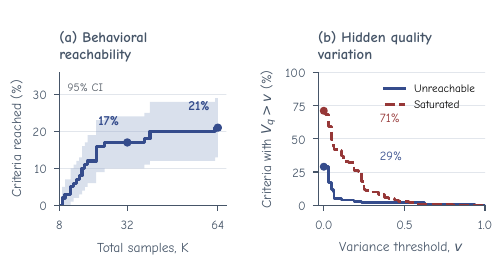}
\caption{\textbf{Probes of inactive rubrics on frozen \mbox{Qwen3-4B}.}
(a) Fraction of initially unreachable rubrics satisfied by at least one of $K$ independent ordinary samples; shading denotes 95\% confidence intervals.
(b) Fraction of initially unreachable and saturated pairs whose estimated quality variation $V_q$ exceeds $v$.}
\label{fig:prelim_probe}
\end{figure}

\paragraph{Rubrics as teachers.}
Rubric guidance recovers most unreachable behaviors but adds no contrast to saturated rubrics. More sampling recovers few: satisfying responses appear for 17\% and 21\% of initially unreachable rubrics at $K=32$ and $64$ (Figure~\ref{fig:prelim_probe}(a)). In contrast, on 100 initially unreachable pairs, appending the target rubric to the prompt raises satisfaction from 1.13\% under fresh ordinary prompting to 91.25\%, versus 3.38\% with generic instructions\footnote{The generic instructions are not length-matched.}. The behavior is therefore within the policy's reach but rarely produced unprompted. For initially saturated rubrics, the same intervention increases parent satisfaction from 98.25\% to 100.00\%, and the judgments stay uniform.

\paragraph{Rubrics as critics.}
Finer evaluation exposes quality differences that saturated judgments hide. An independent judge scores each unchanged baseline response in $[0,1]$ using only the query and parent rubric, without observing binary verdicts, sibling responses, or children. For scores $q_i$, we measure \begin{equation} V_q(x,j)=4\,\operatorname{Var}(q_1,\ldots,q_G). \label{eq:quality_variation} \end{equation} Nonzero variation occurs in 71\% of initially saturated rubrics, compared with 29\% of initially unreachable rubrics (Figure~\ref{fig:prelim_probe}(b)). Child rubrics turn this variation into reward contrast: children selected on baseline responses and evaluated on fresh responses raise the mean held-out contrast from 0.053 for the saturated parent to 0.365. Children also add contrast for unreachable rubrics (from 0.036 to 0.230), but they can only reward partial progress already present in the samples; the target behavior itself still has no positive example. Independent quality scores remain LLM-based diagnostics rather than expert ground truth.

\paragraph{Implications.}

The two endpoints thus call for different roles. An unreachable rubric should guide the policy toward a behavior it can produce but rarely does, and the policy must learn it without seeing the rubric at inference. A saturated rubric should be evaluated at a finer grain that separates current responses, without changing how much it counts in the reward. Section~\ref{sec:method} realizes these roles as Rubric as Teacher and Rubric as Critic.

\section{Method}
\label{sec:method}
\begin{figure*}[t]
\centering
\includegraphics[width=\textwidth]{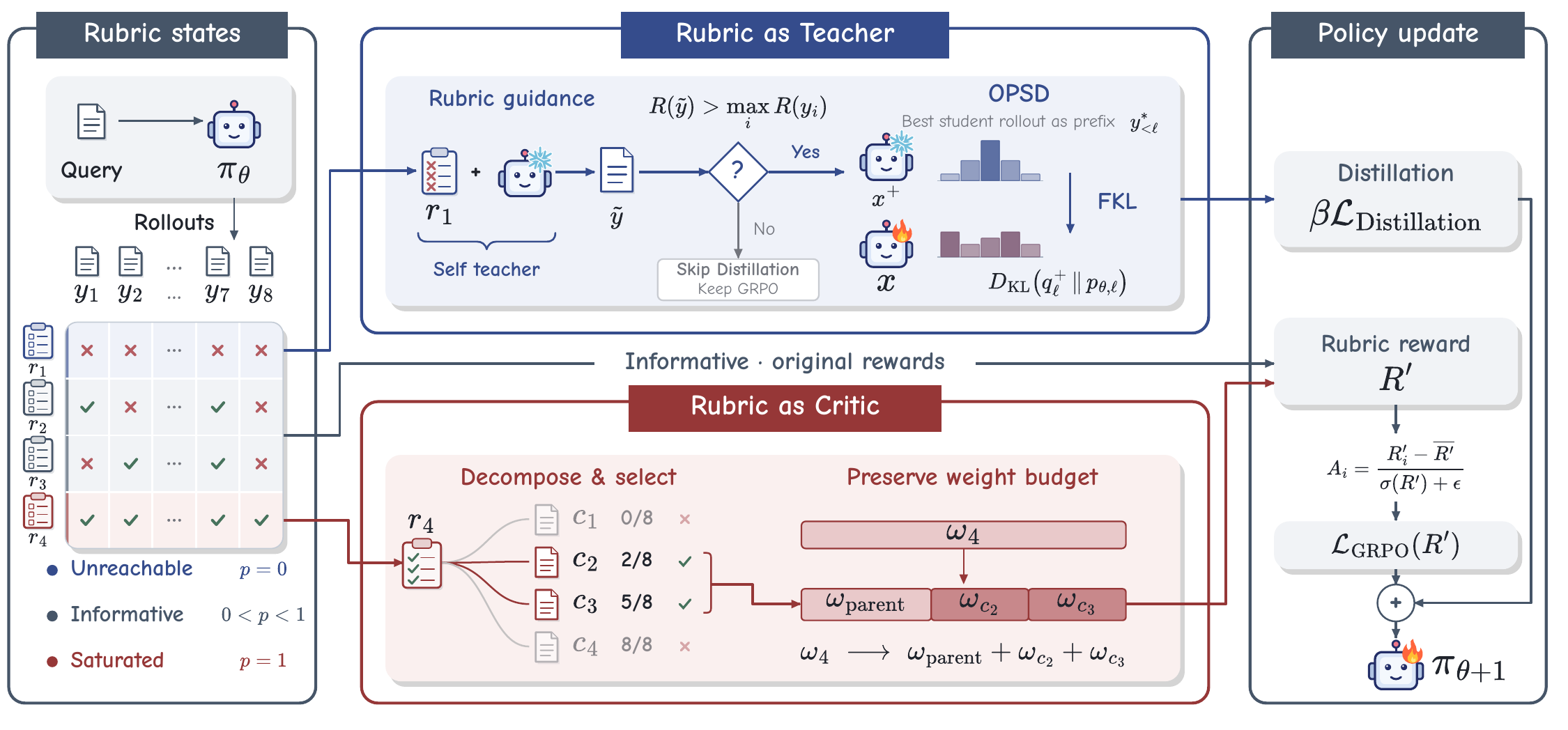}
\caption{\textbf{Rubrics as adaptive critics and teachers.}
Rollout judgments identify unreachable ($p=0$), informative ($0<p<1$), and saturated ($p=1$) rubrics.
As teachers, unreachable rubrics provide privileged guidance, with reward-gated forward-KL supervision on prefixes of the best ordinary rollout.
As critics, saturated rubrics are expanded into discriminative children sharing the parent weight budget.
Original and updated rubric rewards drive GRPO, while teacher supervision contributes an auxiliary objective.}
\label{fig:method}
\end{figure*}

RuACT implements the two roles of Section~\ref{sec:prelim_recovery} and reassigns them in every rollout group as rubric states change (Figure~\ref{fig:method}).
Unreachable rubrics guide a privileged self-teacher to provide behavioral supervision, whereas saturated rubrics are expanded into more discriminative evaluation items for reward construction.
Informative rubrics retain their original rewards.
The two roles jointly optimize the student policy by introducing a distillation loss and reshaping group-relative advantages, respectively.

\subsection{Rubric as Teacher}
\label{sec:teacher}

For unreachable rubrics, ordinary judgments provide no direct within-group contrast.
RuACT instead uses their requirements to condition a frozen self-teacher, the rollout policy $\pi_{\theta_{\rm old}}$, providing behavioral guidance beyond these constant judgments~\citep{hubotter2026sdpo}.
For query $x$, let $\{y_i\}_{i=1}^{G}$ be ordinary responses sampled from $\pi_{\theta_{\rm old}}(\cdot\mid x)$.
We select up to three unreachable rubrics with the largest absolute weights and append them to $x$, yielding a privileged prompt $x^+$.

Following the strategy of CriPO~\citep{xia2026cripo}, we select the highest-reward ordinary response to anchor teacher supervision:
\begin{equation}
i^*=\arg\max_{1\leq i\leq G} R(y_i),
\qquad
y^*=y_{i^*}.
\label{eq:student-prefix}
\end{equation}

Since privileged guidance may improve targeted behaviors at the expense of other requirements~\citep{kim2026selfdistillationdegrade,zhu2026manyfaces,li2026rethinkingopd}, we introduce a \emph{reward-improvement gate} that accepts a rubric-guided response $\tilde{y}$ sampled from $x^+$ by the frozen policy only if its overall reward exceeds that of every ordinary rollout:
\begin{equation}
a_x=
\ind\!\left[
R(\widetilde y)>
\max_{1\leq i\leq G}R(y_i)
\right].
\label{eq:teacher-gate}
\end{equation}

The guided response only gates distillation; supervision stays on the student's own text. For groups passing the reward-improvement gate, teacher and student are evaluated on identical prefixes of $y^*$ but receive different prompt information.
At each supervised position $t$,
\begin{equation}
\begin{aligned}
q_t&=
\operatorname{stopgrad}\!\left[
\pi_{\theta_{\rm old}}(\cdot\mid x^+,y^*_{<t})
\right],\\
p_t&=\pi_{\theta}(\cdot\mid x,y^*_{<t}).
\end{aligned}
\label{eq:teacher-distributions}
\end{equation}
Only the teacher observes the target rubrics.
The ordinary policy receives token-level supervision through forward KL~\citep{agarwal2024gkd,pan2026rlcsd}:
\begin{equation}
\ell_{\rm Distillation}(x)=
\frac{a_x}{|M_x|}
\sum_{t\in M_x}
D_{\rm KL}(q_t\|p_t),
\label{eq:teacher-loss}
\end{equation}
where $M_x$ contains the valid supervised positions.
This encourages the student to follow rubric-conditioned guidance without requiring rubric input at inference.
Further implementation details and the privileged teacher prompt are provided in Appendices~\ref{app:implementation} and~\ref{app:teacher-prompt}, respectively.

\subsection{Rubric as Critic}
\label{sec:critic}

For saturated rubrics, uniform success judgments eliminate within-group reward contrast, potentially masking finer-grained differences in response quality.
RuACT adapts their critic role by evaluating finer aspects of the parent requirement, enabling finer-grained distinctions among responses that receive identical parent-level rewards~\citep{fan2024sedareval,shen2026rrd}.

For a selected saturated parent rubric $r_j$, a
proposal model initially generates candidate child
rubrics using only the query and parent description,
without observing rollout responses.
We evaluate the candidates on current rollouts to
identify those that restore within-group reward
contrast. Let $e_{ijk}\in\{0,1\}$ denote the judgment
of response $i$ under candidate child $k$.
Eligible children must have non-endpoint pass rates:
\begin{equation}
\widetilde{\mathcal K}_j=
\left\{
k: 0<\sum_{i=1}^{G}e_{ijk}<G
\right\}.
\label{eq:childgate}
\end{equation}
When too few candidates qualify, their judgment feedback guides an additional proposal round. We retain at most three eligible children per parent, forming $\mathcal K_j\subseteq\widetilde{\mathcal K}_j$, for reward discrimination.

The parent and its retained children share the
original reward weight budget. For
$m_j=|\mathcal K_j|>0$, their combined contribution is
\begin{equation}
R'_{ij}
=
\frac{w_j}{Z_x}
\left[
(1-\lambda)e_{ij}
+
\frac{\lambda}{m_j}
\sum_{k\in\mathcal K_j}e_{ijk}
\right].
\label{eq:family}
\end{equation}
Here $0<\lambda\leq1$ controls the weight allocated to children. While the parent may remain saturated, its children can introduce additional reward contrast without increasing the family's weight budget.
If no valid child is found, the parent contribution is preserved. We use $\lambda=0.50$. Further implementation details and child-rubric proposal prompts are provided in Appendices~\ref{app:implementation} and~\ref{app:critic-prompt}, respectively.

\subsection{Joint Optimization}
\label{sec:joint}
The critic determines the rubric rewards used to compute group-relative advantages, while the teacher provides an auxiliary token-level objective~\citep{li2026srpo}.
For a batch $\mathcal B$ containing $Q$ query groups, RuACT optimizes

\begin{equation}
\mathcal L_{\mathrm{RuACT}}(\theta)
=
\mathcal L_{\rm GRPO}(\theta;R')
\!+\!
\frac{\beta}{Q}
\sum_{x\in\mathcal B}
\ell_{\rm Distill}(x),
\label{eq:objective}
\end{equation}

where $R'$ combines unchanged rubric rewards with updated parent-family contributions.
Groups rejected by the teacher gate contribute zero auxiliary loss but remain in GRPO, including any valid critic-side reward updates.
We use $\beta=0.1$ for Qwen3-1.7B and $\beta=0.05$ for Qwen3-4B.

\section{Experiments}
\label{sec:experiment}

\begin{table*}[t]

\centering
\begingroup

\fontsize{9.2}{11.2}\selectfont
\setlength{\tabcolsep}{3.5pt}

\renewcommand{\arraystretch}{1.0}
\setlength{\extrarowheight}{0pt}

\definecolor{HeaderGray}{HTML}{EFEFEF}
\definecolor{ResultGain}{HTML}{237A36}
\definecolor{ResultDrop}{HTML}{C91E2B}
\definecolor{ResultZero}{HTML}{999999}
\definecolor{GroupGray}{HTML}{62666F}
\definecolor{OursBlue}{HTML}{285F91}
\definecolor{OursShade}{HTML}{F1F6FA}

\newcommand{\resultgain}[1]{%
  {\fontsize{7.5}{8.5}\selectfont
  \textcolor{ResultGain}{$\uparrow$\,#1}}%
}

\newcommand{\resultdrop}[1]{%
  {\fontsize{7.5}{8.5}\selectfont
  \textcolor{ResultDrop}{$\downarrow$\,#1}}%
}

\newcommand{\resultzero}{%
  {\fontsize{7.5}{8.5}\selectfont
  \textcolor{ResultZero}{$\uparrow$\,0.00}}%
}

\newcommand{\resultcell}[2]{%
  \makebox[2.35em][r]{#1}\,
  \makebox[2.25em][l]{#2}%
}

\newcommand{\reportedcell}[1]{%
  \resultcell{#1}{}%
}

\newcommand{\groupbase}{%
  {\fontsize{7.7}{9}\selectfont
  \textcolor{GroupGray}{\textbf{Vanilla}}}%
}

\newcommand{\groupbaseline}{%
  {\fontsize{7.7}{9}\selectfont
  \textcolor{GroupGray}{\textbf{Baselines}}}%
}

\newcommand{\groupours}{%
  {\fontsize{7.7}{9}\selectfont
  \textcolor{OursBlue}{\textbf{Ours}}}%
}

\begin{tabularx}{\textwidth}{
>{\raggedright\arraybackslash}m{1.78cm}
!{\vrule width 0.4pt}
>{\centering\arraybackslash}m{1.22cm}
!{\vrule width 0.4pt}
>{\raggedright\arraybackslash}m{2.20cm}
!{\vrule width 0.4pt}
>{\centering\arraybackslash}X
>{\centering\arraybackslash}X
!{\vrule width 0.4pt}
>{\centering\arraybackslash}X
>{\centering\arraybackslash}X
!{\vrule width 0.4pt}
>{\centering\arraybackslash}X
}

\toprule[1.0pt]


\rowcolor{HeaderGray}
&
&
&
\multicolumn{2}
{c!{\vrule width 0.4pt}}
{\textbf{Medicine}}
&
\multicolumn{2}
{c!{\vrule width 0.4pt}}
{\textbf{Science}}
&
\\

\rowcolor{HeaderGray}
\multirow{-2}{*}{\textbf{Model}}
&
\multirow{-2}{*}{\textbf{Group}}
&
\multirow{-2}{*}{\textbf{Method}}
&
\textbf{RaR-Med.}
&
\textbf{HealthBench}
&
\textbf{RaR-Sci.}
&
\textbf{ResearchQA}
&
\multirow{-2}{*}{\textbf{Avg.}}
\\

\midrule[0.7pt]



\multirow[c]{8}{*}{\textbf{Qwen3-1.7B}}
& \groupbase
& Base
& \resultcell{38.57}{\resultzero}
& \resultcell{23.18}{\resultzero}
& \resultcell{52.56}{\resultzero}
& \resultcell{50.51}{\resultzero}
& \resultcell{41.21}{\resultzero}
\\

\cmidrule(lr){2-8}


& \multirow[c]{4}{*}{\groupbaseline}
& GRPO
& \resultcell{43.26}{\resultgain{4.69}}
& \resultcell{26.45}{\resultgain{3.27}}
& \resultcell{55.67}{\resultgain{3.11}}
& \resultcell{54.79}{\resultgain{4.28}}
& \resultcell{45.04}{\resultgain{3.84}}
\\

&
& Reweight
& \resultcell{42.60}{\resultgain{4.03}}
& \resultcell{25.61}{\resultgain{2.43}}
& \resultcell{56.77}{\resultgain{4.21}}
& \resultcell{56.11}{\resultgain{5.60}}
& \resultcell{45.27}{\resultgain{4.07}}
\\

&
& RuscaRL
& \resultcell{\underline{43.39}}{\resultgain{4.82}}
& \resultcell{23.95}{\resultgain{0.77}}
& \resultcell{56.41}{\resultgain{3.85}}
& \resultcell{\textbf{59.37}}{\resultgain{8.86}}
& \resultcell{45.78}{\resultgain{4.58}}
\\

&
& OPSD$^{*}$
& \reportedcell{26.30}
& \reportedcell{38.50}
& \reportedcell{39.60}
& \reportedcell{38.40}
& \reportedcell{35.70}
\\

\cmidrule(lr){2-8}


& \multirow[c]{3}{*}{\groupours}
& RuACT-Teacher
& \resultcell{\textbf{43.82}}{\resultgain{5.25}}
& \resultcell{25.16}{\resultgain{1.98}}
& \resultcell{\underline{58.42}}{\resultgain{5.86}}
& \resultcell{57.76}{\resultgain{7.25}}
& \resultcell{46.29}{\resultgain{5.09}}
\\

&
& RuACT-Critic
& \resultcell{43.24}{\resultgain{4.67}}
& \resultcell{\textbf{27.97}}{\resultgain{4.79}}
& \resultcell{58.16}{\resultgain{5.60}}
& \resultcell{58.55}{\resultgain{8.04}}
& \resultcell{\underline{46.98}}{\resultgain{5.77}}
\\

\rowcolor{OursShade}
&
& \textbf{RuACT}
& \resultcell{43.35}{\resultgain{4.78}}
& \resultcell{\underline{27.83}}{\resultgain{4.65}}
& \resultcell{\textbf{58.83}}{\resultgain{6.27}}
& \resultcell{\underline{58.65}}{\resultgain{8.14}}
& \resultcell{\textbf{47.17}}{\resultgain{5.96}}
\\

\midrule[0.7pt]



\multirow[c]{8}{*}{\textbf{Qwen3-4B}}
& \groupbase
& Base
& \resultcell{50.38}{\resultzero}
& \resultcell{35.45}{\resultzero}
& \resultcell{66.16}{\resultzero}
& \resultcell{59.16}{\resultzero}
& \resultcell{52.79}{\resultzero}
\\

\cmidrule(lr){2-8}


& \multirow[c]{4}{*}{\groupbaseline}
& GRPO
& \resultcell{53.55}{\resultgain{3.17}}
& \resultcell{37.62}{\resultgain{2.17}}
& \resultcell{70.08}{\resultgain{3.92}}
& \resultcell{64.82}{\resultgain{5.66}}
& \resultcell{56.52}{\resultgain{3.73}}
\\

&
& Reweight
& \resultcell{55.34}{\resultgain{4.96}}
& \resultcell{38.46}{\resultgain{3.01}}
& \resultcell{70.28}{\resultgain{4.12}}
& \resultcell{65.02}{\resultgain{5.86}}
& \resultcell{57.28}{\resultgain{4.49}}
\\

&
& RuscaRL
& \resultcell{55.36}{\resultgain{4.98}}
& \resultcell{37.10}{\resultgain{1.65}}
& \resultcell{69.38}{\resultgain{3.22}}
& \resultcell{65.33}{\resultgain{6.17}}
& \resultcell{56.79}{\resultgain{4.01}}
\\

&
& OPSD$^{*}$
& \reportedcell{42.00}
& \reportedcell{42.10}
& \reportedcell{38.30}
& \reportedcell{45.20}
& \reportedcell{41.90}
\\

\cmidrule(lr){2-8}


& \multirow[c]{3}{*}{\groupours}
& RuACT-Teacher
& \resultcell{55.15}{\resultgain{4.77}}
& \resultcell{38.57}{\resultgain{3.12}}
& \resultcell{\underline{70.53}}{\resultgain{4.37}}
& \resultcell{\underline{66.12}}{\resultgain{6.96}}
& \resultcell{\underline{57.59}}{\resultgain{4.81}}
\\

&
& RuACT-Critic
& \resultcell{\underline{55.50}}{\resultgain{5.12}}
& \resultcell{\underline{38.77}}{\resultgain{3.32}}
& \resultcell{70.47}{\resultgain{4.31}}
& \resultcell{64.62}{\resultgain{5.46}}
& \resultcell{57.34}{\resultgain{4.55}}
\\

\rowcolor{OursShade}
&
& \textbf{RuACT}
& \resultcell{\textbf{56.34}}{\resultgain{5.96}}
& \resultcell{\textbf{40.12}}{\resultgain{4.67}}
& \resultcell{\textbf{70.59}}{\resultgain{4.43}}
& \resultcell{\textbf{69.44}}{\resultgain{10.28}}
& \resultcell{\textbf{59.12}}{\resultgain{6.34}}
\\

\bottomrule[1.0pt]

\end{tabularx}

\endgroup
\caption{\textbf{Main results across medical and scientific reasoning benchmarks.}
Scores are reported in \% on a normalized 0--100 scale; Avg. is the unweighted mean across the four benchmarks.
Arrows indicate absolute score changes relative to the corresponding vanilla checkpoint.
Bold and underlined values denote the best and second-best results
within each model scale, respectively.
OPSD$^{*}$ results are reported from CriPO \citep{xia2026cripo} using GPT-4o-mini
as the judge.}
\label{tab:main-results}
\end{table*}

\subsection{Setup}
\paragraph{Datasets and Evaluation.}
We evaluate RuACT on both medical and scientific reasoning. RaR-Medicine and RaR-Science \citep{gunjal2026rubrics} serve as the training datasets and in-domain evaluation benchmarks for the two domains, respectively. For out-of-domain evaluation, we use HealthBench \citep{arora2025healthbench} for medicine and ResearchQA \citep{yifei2026researchqa} for science. We conduct experiments with Qwen3-1.7B and Qwen3-4B \citep{yang2025qwen3} using a verl-based implementation \citep{sheng2024hybridflow}. All methods are trained for 200 optimization steps under matched budgets, using Qwen3-32B \citep{yang2025qwen3} as the rubric judge during training and Deepseek-v4.1-flash \citep{deepseekai2026deepseekv41flash} as the evaluation judge.  Further details on datasets, training configurations, and evaluation protocols are provided in Appendix~\ref{app:experimental-setup}, and Appendix~\ref{app:judge-robustness} re-evaluates all models with a second judge. 

\paragraph{Baselines and Models.}
We compare RuACT against representative baselines covering different optimization strategies.
\textbf{GRPO} \citep{shao2024deepseekmath} provides standard group-relative optimization, while \textbf{OPSD} \citep{zhao2026opsd} introduces privileged-information-guided distillation.
For rubric-based methods, \textbf{RuscaRL} \citep{zhou2025ruscarl} represents \emph{rubrics as teachers}, guiding policy exploration through rubric-based scaffolding.
In contrast, \textbf{Reweight}, our implementation inspired by POW3R \citep{tyagi2026pow3r}, represents \emph{rubrics as critics}, adaptively weighting rubrics based on their current informativeness.
We additionally evaluate \textbf{RuACT-Teacher} and \textbf{RuACT-Critic} to isolate the contributions of recovering unreachable and saturated rubrics, respectively, and compare with DAPO-style sampling and GRPO+OPSD in Section~\ref{sec:analysis}.
All locally implemented methods share the same training and evaluation protocol (Appendix~\ref{app:experimental-setup}), while OPSD results are quoted from CriPO \citep{xia2026cripo} with another judge.

\subsection{Main Results}
\label{sec:main-results}

\paragraph{Overall Performance.}
As shown in Table~\ref{tab:main-results}, RuACT achieves the best average at both scales, raising GRPO from 45.04\% to 47.17\% on Qwen3-1.7B and from 56.52\% to 59.12\% on Qwen3-4B, and exceeding the strongest local baseline (RuscaRL at 45.78\% and Reweight at 57.28\%, respectively). At 4B, RuACT is best on all four benchmarks, with the largest gains on the out-of-domain HealthBench (from 37.62\% to 40.12\%) and ResearchQA (from 64.82\% to 69.44\%). At 1.7B, the single-role variants already lead on individual benchmarks, and combining them mainly improves the average. Both single-role variants outperform GRPO on average at both scales, so each role contributes independently. Although trained only on medical questions, RuACT also improves over the base model on MT-Bench and FollowBench (Table~\ref{tab:ood-generalization}).

\begin{figure}[t]
    \centering
    \includegraphics[width=\columnwidth]{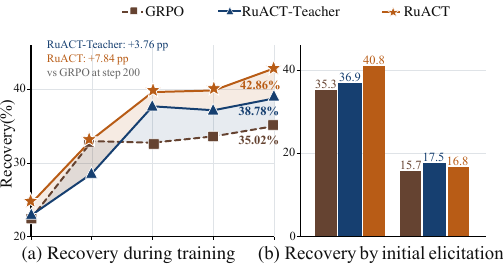}
    \caption{\textbf{Recovery of initially unreachable rubrics} (Qwen3-4B, RaR-Medicine). A rubric is recovered if any of eight unguided rollouts satisfies it. (a) During training. (b) Grouped by whether rubric guidance elicits the behavior from the base model.}
    \label{fig:unreachable-recovery}
\end{figure}

\subsection{Analysis}
\label{sec:analysis}

\paragraph{Unreachable Does Not Mean Unlearnable.}
Rubrics unreachable under the current policy can become learnable through targeted teacher supervision.
We track 931 initially unreachable rubrics from 331 fixed RaR-Medicine training queries, measuring recovery every 40 steps using eight unguided rollouts.
As shown in Figure~\ref{fig:unreachable-recovery}, RuACT-Teacher alone raises the recovery rate at step 200 from 35.02\% under GRPO to 38.78\%, and the full method reaches 42.86\%. Because recovery is measured without privileged prompts, these gains reflect the policy's own generation.
Figure~\ref{fig:unreachable-recovery}(b) shows that RuACT achieves larger average recovery gains on initially rubric-elicitable behaviors (+5.5 pp) than on those not elicited (+1.1 pp).

\begin{table}[t]

\centering
\begingroup

\fontsize{8.3}{10}\selectfont
\setlength{\tabcolsep}{4pt}
\renewcommand{\arraystretch}{1.0}
\setlength{\extrarowheight}{0pt}

\definecolor{HeaderGray}{HTML}{EFEFEF}
\definecolor{OursShade}{HTML}{F1F6FA}

\begin{tabularx}{\columnwidth}{
@{}
>{\raggedright\arraybackslash}m{3.10cm}
!{\vrule width 0.4pt}
>{\centering\arraybackslash}X
!{\vrule width 0.4pt}
>{\centering\arraybackslash}X
@{}
}

\toprule[1.0pt]

\rowcolor{HeaderGray}
\textbf{Scoring rule}
&
\shortstack{
\textbf{Quality}\\
\textbf{alignment} (\%) $\uparrow$
}
&
\shortstack{
\textbf{Reward}\\
\textbf{coverage} (\%) $\uparrow$
}
\\

\midrule[0.7pt]

parent rubric-only
& 57.74
& 65.92
\\

\rowcolor{OursShade}
\textbf{w/ Decomposed}
& \textbf{60.77}
& 88.85
\\

w/ Shuffled Decomposed
& 53.62
& \textbf{90.44}
\\

\bottomrule[1.0pt]

\end{tabularx}

\endgroup
\caption{\textbf{Quality alignment of refined rewards.}
Reward-quality alignment and coverage.}
\label{tab:quality-alignment}
\end{table}

\begin{figure}[t]
    \centering
    \includegraphics[width=\columnwidth]{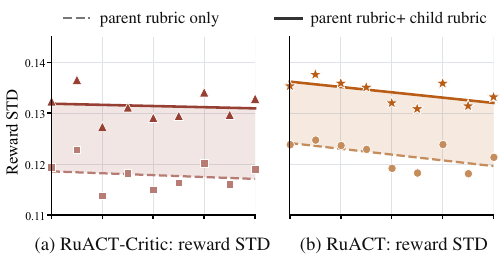}
    \caption{\textbf{Reward discrimination on independent rollouts.}
    Within-group reward STD with and without child-rubric refinement for (a) RuACT-Critic and (b) RuACT on Qwen3-4B RaR-Medicine.
    Rollouts are sampled independently of those used for child selection.}
    \label{fig:refinement-reward-std}
\end{figure}



\paragraph{Saturated Does Not Mean Indistinguishable.}
We evaluate child refinement on fresh rollouts, separate from those used for selection.
Refinement raises within-group reward STD by 11.34\% for RuACT-Critic and 10.02\% for RuACT (Figure~\ref{fig:refinement-reward-std}).

We further assess whether this additional contrast reflects response quality using blind judgments on 300 rollout groups (2,400 responses).
We measure reward coverage as the fraction of response pairs assigned different rewards, and quality alignment as the agreement between reward rankings and independent quality judgments, excluding judge ties and assigning half credit to reward ties.
Both metrics are averaged within groups and then across groups.
Table~\ref{tab:quality-alignment} shows that matched children improve quality alignment from 57.74\% to 60.77\% and reward coverage from 65.92\% to 88.85\%.
Shuffled children achieve higher coverage (90.44\%) but lower alignment (53.62\%), indicating that reward discrimination alone does not guarantee alignment with response quality.

\begin{table}[t]

\centering
\begingroup

\fontsize{8.3}{10}\selectfont
\setlength{\tabcolsep}{2.5pt}
\renewcommand{\arraystretch}{1.0}
\setlength{\extrarowheight}{0pt}

\definecolor{HeaderGray}{HTML}{EFEFEF}
\definecolor{ResultGain}{HTML}{237A36}
\definecolor{ResultZero}{HTML}{999999}
\definecolor{OursShade}{HTML}{F1F6FA}

\newcommand{\resultgain}[1]{%
  {\fontsize{7.0}{8.5}\selectfont
  \textcolor{ResultGain}{$\uparrow$\,#1}}%
}

\newcommand{\resultzero}{%
  {\fontsize{7.0}{8.5}\selectfont
  \textcolor{ResultZero}{$\uparrow$\,0.00}}%
}

\newcommand{\resultcell}[2]{%
  \mbox{#1\hspace{0.25em}#2}%
}

\newcommand{\reportedcell}[1]{%
  \mbox{#1}%
}

\begin{tabularx}{\columnwidth}{
@{}
>{\raggedright\arraybackslash}m{2.10cm}
!{\vrule width 0.4pt}
>{\centering\arraybackslash}X
>{\centering\arraybackslash}X
!{\vrule width 0.4pt}
>{\centering\arraybackslash}X
@{}
}

\toprule[1.0pt]

\rowcolor{HeaderGray}
\textbf{Method}
&
\textbf{RaR-Med.}
&
\textbf{HealthBench}
&
\textbf{Avg.}
\\

\midrule[0.7pt]


Base
& \resultcell{50.38}{\resultzero}
& \resultcell{35.45}{\resultzero}
& \resultcell{42.92}{\resultzero}
\\

\cmidrule(lr){1-4}


GRPO
& \resultcell{53.55}{\resultgain{3.17}}
& \resultcell{37.62}{\resultgain{2.17}}
& \resultcell{45.59}{\resultgain{2.67}}
\\

DAPO-style
& \resultcell{\underline{56.29}}{\resultgain{5.91}}
& \resultcell{\underline{39.27}}{\resultgain{3.82}}
& \resultcell{\underline{47.78}}{\resultgain{4.87}}
\\

GRPO+OPSD
& \resultcell{55.59}{\resultgain{5.21}}
& \resultcell{38.63}{\resultgain{3.18}}
& \resultcell{47.11}{\resultgain{4.20}}
\\

OPSD$^{*}$
& \reportedcell{42.00}
& \reportedcell{42.10}
& \reportedcell{42.05}
\\

\midrule[0.7pt]


\rowcolor{OursShade}
\textbf{RuACT}
& \resultcell{\textbf{56.34}}{\resultgain{5.96}}
& \resultcell{\textbf{40.12}}{\resultgain{4.67}}
& \resultcell{\textbf{48.23}}{\resultgain{5.32}}
\\

\bottomrule[1.0pt]

\end{tabularx}

\endgroup
\caption{\textbf{Comparison with alternative optimization strategies on Qwen3-4B.}
Scores are reported in \%; arrows indicate gains over Base.
DAPO-style adapts dynamic sampling to rubric-based GRPO~\citep{yu2025dapo}.}
\label{tab:analysis}
\end{table}

\begin{figure}[t]
    \centering
    \includegraphics[width=\columnwidth]{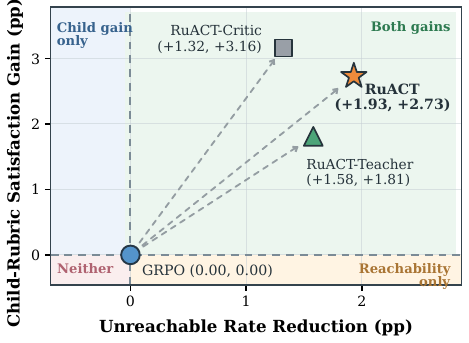}
    \caption{\textbf{Policy-level gains across inactive rubrics.}
    The x-axis measures unreachable-rate reduction over 21,500 test query--rubric pairs.
    The y-axis measures child-rubric satisfaction gains on 1,954 fixed Base-saturated parents with shared children.
    Both axes show percentage-point improvements relative to GRPO.}
    \label{fig:capability-map}
\end{figure}

\paragraph{Joint Improvements across Inactive Rubrics.}
We examine how the teacher and critic roles jointly improve policy behavior.
Figure~\ref{fig:capability-map} compares unreachable-rate reduction and child-rubric satisfaction gains relative to GRPO.
RuACT-Teacher achieves a larger unreachable reduction than RuACT-Critic (+1.58 vs.\ +1.32 pp), while RuACT-Critic yields greater fine-grained satisfaction gains (+3.16 vs.\ +1.81 pp).
RuACT achieves the largest unreachable reduction (+1.93 pp) while maintaining substantial child-rubric satisfaction gains (+2.73 pp).
These results show the distinct contributions of both roles and their joint benefits.

\paragraph{Comparison with Alternative Optimization Strategies.}
We further compare RuACT with dynamic sampling and privileged-information distillation strategies.
As shown in Table~\ref{tab:analysis}, RuACT achieves the highest average among locally evaluated methods (48.23\%), above DAPO-style sampling (47.78\%) and GRPO+OPSD (47.11\%). DAPO-style sampling nearly matches RuACT on RaR-Medicine (56.29\% vs.\ 56.34\%) but trails on HealthBench (39.27\% vs.\ 40.12\%). The two act at different levels: dynamic sampling discards groups whose total rewards are constant, whereas RuACT recovers individual rubrics inside groups whose total rewards still vary, so they can be combined.


\begin{table}[t]

\centering
\begingroup

\fontsize{8.2}{10}\selectfont
\setlength{\tabcolsep}{2.5pt}
\renewcommand{\arraystretch}{1.0}
\setlength{\extrarowheight}{0pt}

\definecolor{HeaderGray}{HTML}{EFEFEF}
\definecolor{OursShade}{HTML}{F1F6FA}

\begin{tabularx}{\columnwidth}{
@{}
>{\raggedright\arraybackslash}m{1.65cm}
!{\vrule width 0.4pt}
>{\raggedright\arraybackslash}m{1.10cm}
!{\vrule width 0.4pt}
>{\centering\arraybackslash}X
>{\centering\arraybackslash}X
!{\vrule width 0.4pt}
>{\centering\arraybackslash}X
>{\centering\arraybackslash}X
@{}
}

\toprule[1.0pt]


\rowcolor{HeaderGray}
&
&
\multicolumn{2}{c!{\vrule width 0.4pt}}
{\textbf{MT-Bench}}
&
\multicolumn{2}{c}
{\textbf{FollowBench}}
\\

\rowcolor{HeaderGray}
\multirow{-2}{*}{\textbf{Model}}
&
\multirow{-2}{*}{\textbf{Method}}
&
\textbf{R1}
&
\textbf{R2}
&
\textbf{HSR}
&
\textbf{SSR}
\\

\midrule[0.7pt]


Qwen3-1.7B
& Base
& 7.13
& 5.90
& 43.52
& 56.44
\\

\rowcolor{OursShade}
& \textbf{RuACT}
& \textbf{7.66}
& \textbf{6.03}
& \textbf{45.37}
& \textbf{57.69}
\\

\midrule[0.7pt]


Qwen3-4B
& Base
& 8.31
& 7.60
& 58.50
& 67.28
\\

\rowcolor{OursShade}
& \textbf{RuACT}
& \textbf{8.64}
& \textbf{7.88}
& \textbf{59.89}
& \textbf{67.75}
\\

\bottomrule[1.0pt]

\end{tabularx}

\endgroup
\caption{\textbf{General capability preservation on OOD benchmarks.}
MT-Bench scores (0--10) and FollowBench success rates (\%) are evaluated using Qwen3-Max as the judge.}
\label{tab:ood-generalization}
\end{table}

\paragraph{General Capability Preservation}
We further examine whether medicine-specific training compromises general instruction-following capabilities.
Using the RaR-Medicine-trained checkpoints, we evaluate Base and RuACT on MT-Bench and FollowBench with Qwen3-Max as the judge.
As shown in Table~\ref{tab:ood-generalization}, RuACT improves both MT-Bench turns and FollowBench success rates at both model scales.
These results suggest that recovering inactive supervision preserves the evaluated general capabilities while improving domain-specific performance.

\section{Ablation Study}

\begin{table}[t]

\centering
\begingroup

\fontsize{8.2}{10}\selectfont
\setlength{\tabcolsep}{3pt}
\renewcommand{\arraystretch}{1.0}
\setlength{\extrarowheight}{0pt}

\definecolor{HeaderGray}{HTML}{EFEFEF}
\definecolor{OursShade}{HTML}{F1F6FA}

\begin{tabularx}{\columnwidth}{
@{}
>{\raggedright\arraybackslash}X
>{\centering\arraybackslash}c
>{\centering\arraybackslash}c
>{\centering\arraybackslash}c
@{}
}

\toprule[1.0pt]

\rowcolor{HeaderGray}
\textbf{Variant}
& \textbf{RaR-Med.}
& \textbf{HealthBench}
& \textbf{Avg.}
\\

\midrule[0.8pt]


\rowcolor{OursShade}
\textbf{RuACT}
& \textbf{56.34}
& \textbf{40.12}
& \textbf{48.23}
\\

\midrule[0.7pt]


\multicolumn{4}{@{}l}{
\textbf{\textit{Rubric as Teacher}}
}
\\

\hspace{0.7em}RuACT-Teacher
& 55.15
& 38.57
& 46.86
\\

\shortstack[l]{
\hspace{0.7em}w/o Gate
}
& 55.18
& 37.45
& 46.32
\\

\midrule[0.7pt]


\multicolumn{4}{@{}l}{
\textbf{\textit{Rubric as Critic}}
}
\\

\hspace{0.7em}RuACT-Critic
& 55.50
& 38.77
& 47.14
\\

\shortstack[l]{
\hspace{0.7em}w/o Filter
}
& 54.99
& 38.50
& 46.75
\\

\bottomrule[1.0pt]

\end{tabularx}

\endgroup
\caption{\textbf{Ablation study of RuACT on Qwen3-4B.}
Scores are reported in \% on RaR-Medicine and HealthBench, with
Avg. denoting their unweighted mean.}
\label{tab:ablation}
\end{table}

\paragraph{Both Recovery Roles Matter.}
Neither teacher guidance nor critic refinement alone matches the full RuACT framework.
As shown in Table~\ref{tab:ablation}, RuACT achieves an average of 48.23\%, compared with 46.86\% for RuACT-Teacher and 47.14\% for RuACT-Critic.
This supports the distinct contributions of recovering unreachable behaviors and restoring reward discrimination.

\paragraph{Teacher Guidance Must Be Selective.}
Rubric-conditioned guidance does not always yield better overall responses.
Across 10,780 gate evaluations in full RuACT training, only 44.86\% of guided candidates exceed the best student rollout in aggregate rubric reward, and the acceptance rate decreases from 50.69\% in the first 50 steps to 42.61\% in the last 50 steps, suggesting fewer opportunities for reward-improving guidance later in training.
Removing the gate leaves RaR-Medicine nearly unchanged (55.15\% vs.\ 55.18\%) but reduces HealthBench from 38.57\% to 37.45\%, lowering the average from 46.86\% to 46.32\%.
The gate thus blocks guided responses that improve the target rubrics at the expense of others.

\paragraph{Effective Refinement Requires Filtering.}
Generating child rubrics does not guarantee informative reward distinctions.
Among 93,523 judged child candidates, 64.10\% receive uniform judgments across all eight rollouts, while only 21.06\% are ultimately retained after filtering and selection limits.
This limits unnecessary reward modifications from candidates without within-group contrast.
Moreover, the proportion of fully satisfied candidates increases from 43.49\% to 54.00\% across training, highlighting the continued need for filtering.
Removing the filter reduces the average from 47.14\% to 46.75\%, indicating that only children that separate current responses are useful.

\section{Related Work}
\label{sec:related_work}

\paragraph{Rubrics as rewards and critics.}
Rubric-based RL aggregates judgments on checkable criteria into rewards~\citep{gunjal2026rubrics,huang2025rubricanchors,viswanathan2025checklists,dineen2025qalign,he2026advancedif}, using expert-written or LLM-synthesized rubrics~\citep{li2026ares,liu2026openrubrics,shen2026rrd,zhang2026rubricbench}. Adaptive critics reweight or schedule rubrics~\citep{tyagi2026pow3r,chen2026rucl}, reshape how judgments become rewards~\citep{xu2026rubricarm,lan2026arlrr,xu2026rubricstotokens,zhang2026cort}, or evolve rubrics with the policy~\citep{shao2025drtulu,rezaei2025onlinerubrics,ding2026evorubrics,wang2026dynamicrubric,sheng2026rlcer,xu2026sibylsense}. Reweighting cannot fix a constant rubric, and evolving methods revise the whole rubric set; RuACT refines only saturated rubrics within their weight.

\paragraph{Rubrics as teachers.}
Rubrics also guide generation through scaffolded exploration or hindsight replay~\citep{zhou2025ruscarl,zhang2026herl}, or serve as privileged context distilled into the policy~\citep{zhao2026opsd,hubotter2026sdpo,ye2026opcd,rezaei2026rgsd,gu2026rcsd,pan2026rlcsd}, or combined with group-relative updates~\citep{li2026srpo,han2026rstg,lu2026sdar}. Distillation supplies dense token-level supervision~\citep{agarwal2024gkd}, but privileged teachers may also transfer undesirable behaviors~\citep{kim2026selfdistillationdegrade,zhu2026manyfaces,li2026rethinkingopd}, so guidance should be verified before distillation. Closest to ours, CriPO injects unmet criteria into a teacher anchored on the best student response and corrects advantages for suppressed criteria~\citep{xia2026cripo}. These methods cover only the failing side. RuACT teaches only rubrics all rollouts fail, reward-gates the guidance, and adds a critic for saturated ones.
Unlike direct imitation of guided responses, RuACT distills rubric-conditioned distributions on prefixes of ordinary student rollouts.

\section{Conclusion}
\label{sec:conclusion}
Rubric inactivity does not necessarily imply a lack of learning value.
We introduce \textbf{RuACT}, which adaptively uses rubrics as teachers and critics to recover supervision lost under the current policy.
For unreachable rubrics, reward-gated teacher guidance helps the policy acquire previously unmet behaviors; for saturated rubrics, critic refinement restores fine-grained reward discrimination.
Across medical and scientific reasoning benchmarks, RuACT achieves the highest average performance at both model scales.
Analyses show improved unguided recovery of unreachable rubrics and more quality-aligned reward signals from refined rubrics.
These findings highlight the value of recovering behavioral and evaluative supervision.

\section*{Limitations}
RuACT relies on LLM-based rubric judgments and child-rubric proposals. Although our experiments evaluate reward--quality alignment using an independent judge, the generated children may not always faithfully capture the intended requirements. Improving the reliability of rubric refinement remains a direction for future work.

Rubric states are identified from a limited group of $G=8$ responses. Consequently, unreachable and saturated labels reflect the current samples rather than intrinsic properties of the policy. Larger or adaptive groups could improve state estimation, especially for rarely satisfied rubrics.

Our experiments primarily cover medicine and science with Qwen3-1.7B and Qwen3-4B. Generalization to larger models and broader domains remains to be explored. Moreover, teacher guidance and critic refinement require additional inference during training, which future work could reduce.

\section*{Ethical Statement}
Our models are trained and evaluated on medical and scientific question answering. All evaluations rely on LLM judges rather than expert review, and the trained models are research artifacts that should not be used for clinical advice or decisions without professional oversight. Rubric-guided distillation may also propagate biases present in the rubrics or judges. We use public datasets and models.

\bibliography{references}

\clearpage
\appendix

\section{Detailed Experimental Settings}
\label{app:experimental-setup}

\subsection{Datasets}
\label{app:datasets}

We train on RaR-Medicine and RaR-Science~\citep{gunjal2026rubrics} and evaluate on their test splits, HealthBench~\citep{arora2025healthbench}, and ResearchQA~\citep{yifei2026researchqa}. Table~\ref{tab:app-datasets} reports the sizes of the prepared inputs. HealthBench and ResearchQA each use a fixed 500-example subset for evaluation only. The HealthBench subset is sampled with seed 42. All methods evaluated in our experiments use the same frozen evaluation inputs and rubric weights.

\begin{table}[htbp]
\centering
\caption{Dataset sizes used in our experiments. Counts refer to examples in the prepared splits. HealthBench and ResearchQA are evaluation subsets.}
\label{tab:app-datasets}
\small
\setlength{\tabcolsep}{5pt}
\begin{tabular*}{\columnwidth}{@{\extracolsep{\fill}}lrr@{}}
\toprule
Dataset & Train & Evaluation \\
\midrule
RaR-Medicine & 17,926 & 2,242 \\
RaR-Science & 18,302 & 2,292 \\
HealthBench & -- & 500 \\
ResearchQA & -- & 500 \\
\bottomrule
\end{tabular*}
\end{table}

\subsection{Training Setup}
\label{app:training-config}

Policies are initialized from Qwen3-1.7B or Qwen3-4B and trained with \texttt{verl}, BF16 FSDP, vLLM, and AdamW on two NVIDIA A800 GPUs. Table~\ref{tab:app-training-config} gives the shared optimization settings and the standard RaR-Medicine sequence limits. Methods use the same initialization, training data, seed, query batch size, and rollout group size unless a difference is specified below. We match the number of training steps. Teacher guidance and rubric generation incur additional inference cost.

\begin{table}[htbp]
\centering
\caption{Policy-training settings. Sequence limits are for the standard RaR-Medicine configuration.}
\label{tab:app-training-config}
\small
\setlength{\tabcolsep}{5pt}
\renewcommand{\arraystretch}{1.08}
\begin{tabularx}{\columnwidth}{@{}>{\raggedright\arraybackslash}Xr@{}}
\toprule
Hyperparameter & Value \\
\midrule
Training steps & 200 \\
Queries per batch & 64 \\
Responses per query, $G$ & 8 \\
PPO mini-batch size & 32 \\
PPO epochs per batch & 1 \\
Learning rate & $10^{-6}$ \\
Learning-rate warmup steps & 12 \\
Reference KL coefficient & $10^{-3}$ \\
Rollout temperature & 1.0 \\
Rollout top-$p$ & 1.0 \\
Maximum prompt length & 1,024 tokens \\
Maximum response length & 4,096 tokens \\
Maximum context length & 6,144 tokens \\
\bottomrule
\end{tabularx}
\end{table}

We use seed 42, disable thinking and top-$k$ filtering, and save checkpoints every 20 steps. The actor micro-batch size is one per GPU, except for Reweight and RuscaRL, which use four. Dynamic sampling is enabled only in the DAPO-style baseline.

\subsection{Baseline Implementations}
\label{app:baseline-details}

\paragraph{GRPO.}
GRPO~\citep{shao2024deepseekmath} optimizes the original rubric reward without teacher guidance or child-rubric refinement.

\paragraph{Reweight.}
We implement a group-based reweighting control inspired by POW3R~\citep{tyagi2026pow3r}. It recomputes weights from the current response group, without maintaining a historical weighting state. Let $g_{ij}$ denote binary satisfaction after polarity adjustment: for a negative-weight criterion, $g_{ij}$ is one minus its raw satisfaction verdict. For responses with valid judgments, indexed by $\mathcal G_j$, we compute
\begin{equation}
\begin{aligned}
\widehat p_j &= \frac{1}{|\mathcal G_j|}\sum_{i\in\mathcal G_j}g_{ij},\\
a_j &= 1+4\eta\widehat p_j(1-\widehat p_j),
\end{aligned}
\label{eq:app-reweight-factor}
\end{equation}
where $\eta=0.5$. With $\widetilde w_j=|w_j|a_j$, the response reward is
\begin{equation}
R_i^{\mathrm{reweight}}=
\frac{\sum_{j\in\mathcal J_i}\widetilde w_jg_{ij}}
     {\sum_{j\in\mathcal J_i}\widetilde w_j},
\label{eq:app-reweight-reward}
\end{equation}
where $\mathcal J_i$ contains valid judgments for response $i$. Missing verdicts are excluded. Uniformly passed or failed criteria retain their original weight magnitude. This baseline implements the variance-based reweighting idea rather than the full POW3R algorithm.

\paragraph{RuscaRL.}
Our RuscaRL~\citep{zhou2025ruscarl} implementation generates rubric-guided responses and uses the vanilla policy loss. The Qwen3-4B RaR-Science run uses prompt and context limits of 8,192 and 12,288 tokens, respectively, with vLLM memory utilization set to 0.50.

\paragraph{GRPO+OPSD.}
This baseline augments GRPO with distillation from a teacher conditioned on the complete rubric. It does not use unreachable-rubric selection or critic refinement. Our gated variant accepts guidance using the original rubric reward and uses an ordinary rollout for distillation. For Qwen3-4B, the effective teacher coefficient is 0.05 with a 50-step warmup.

\paragraph{DAPO-style Sampling.}
This baseline adds bounded dynamic sampling to the GRPO objective. Other components of DAPO are not included.

\subsection{RuACT Implementation}
\label{app:implementation}

Table~\ref{tab:app-ruact-hparams} lists the teacher and critic settings. The objectives and child-eligibility rule are defined in Section~\ref{sec:method}; prompts are provided in Appendix~\ref{app:prompts}.

\begin{table}[htbp]
\centering
\caption{RuACT hyperparameters. Critic settings describe the Qwen3-4B RaR-Medicine configuration. Selection and retention counts are maxima.}
\label{tab:app-ruact-hparams}
\small
\setlength{\tabcolsep}{5pt}
\renewcommand{\arraystretch}{1.08}
\begin{tabularx}{\columnwidth}{@{}>{\raggedright\arraybackslash}Xr@{}}
\toprule
Hyperparameter & Setting \\
\midrule
\multicolumn{2}{@{}l}{\textit{Teacher}} \\
Target rubrics per query & 3 \\
Guided responses per selected query & 1 \\
Effective $\beta$ (1.7B / 4B) & 0.10 / 0.05 \\
Coefficient warmup steps & 50 \\
\addlinespace[4pt]
\multicolumn{2}{@{}l}{\textit{Critic}} \\
Selected parents per query & 3 \\
Initial candidates per parent & 5 \\
Additional proposal rounds (max.) & 1 \\
Retained children per parent & 3 \\
Active children per query & 3 \\
Parent--child mixture, $\lambda$ & 0.50 \\
\bottomrule
\end{tabularx}
\end{table}

\paragraph{Teacher.}
We select unreachable rubrics by descending absolute weight and generate one guided response per selected query. The full original-rubric reward determines gate acceptance. For accepted groups, distillation uses the highest-reward valid ordinary rollout as its trajectory, with the teacher and student conditioned on the same response prefix. Only the teacher receives the privileged rubric instructions. Rejected groups retain the GRPO update. Forward KL uses the top-64 vocabulary entries and a residual bin. The reported $\beta$ is the effective coefficient, equal to the internal loss coefficient multiplied by $G=8$.

\paragraph{Critic.}
Candidates are proposed from the query and parent description. Rollout texts are used for validation and selection, not proposal generation. If too few candidates pass, one further proposal round uses feedback on rejected candidates. Up to three children are retained per selected parent, while at most three children per query enter the active reward view. Thus, in Eq.~\ref{eq:family}, $m_j$ counts the active children of parent $j$. A parent with no active child keeps its original contribution. Original GRPO rubrics are not subject to the child cap.

\subsection{Evaluation and Judges}
\label{app:evaluation-protocols}

Qwen3-32B supplies binary judgments during training, including child-rubric validation. Qwen3.6-35B-A3B generates child-rubric proposals. We use DeepSeek-V4.1-Flash for the main benchmark evaluation and Qwen3.7-Max for the judge-robustness evaluation.

For each evaluation query, we generate one response with temperature 0.70, top-$p$ 0.80, seed 42, and a 4,096-token response limit. Thinking and top-$k$ filtering are disabled. The same frozen responses are used across evaluation judges. RaR scores use binary judgments and the explicit signed rubric weights. HealthBench divides signed rubric points by the total positive weight. Aggregation is fixed across methods within each benchmark.

OPSD$^{*}$ scores quoted from CriPO~\citep{xia2026cripo} are external reference results. CriPO uses filtered RaR splits and a different evaluation judge, so these scores are reported separately from our matched comparisons.

\section{Supplementary Training Analyses}
\subsection{Rubric Inactivity across Training}
\label{app:inactivity-statistics}

\begin{figure}[t]
    \centering
    \includegraphics[
        width=\columnwidth,
    ]{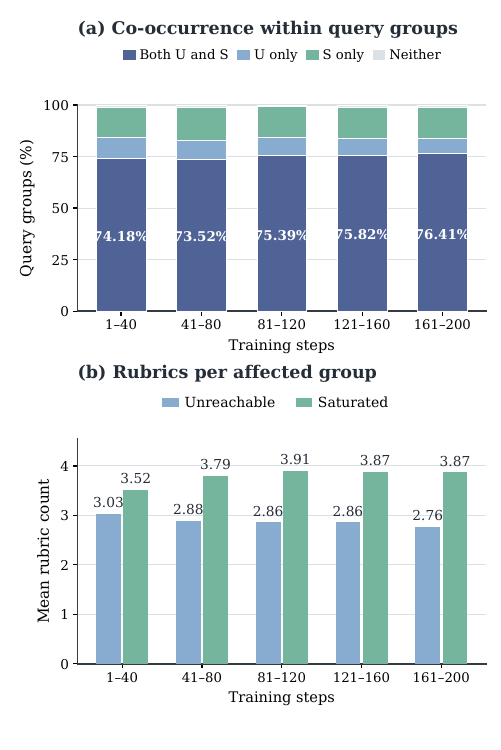}
    \caption{\textbf{Rubric inactivity persists and
    co-occurs throughout GRPO training.}
    (a) Query-group composition by unreachable (U)
    and saturated (S) rubric occurrence across five
    40-step phases.
    (b) Mean number of U or S rubrics per affected
    group, conditional on at least one occurrence.
    Statistics use the original rubrics and eight
    rollouts per query group from Qwen3-4B training
    on RaR-Medicine.}
    \label{fig:inactivity-across-training}
\end{figure}

\paragraph{Both Forms of Inactivity Persist and Co-occur.}
We examine rubric inactivity across all 200 steps of
Qwen3-4B GRPO training on RaR-Medicine, comprising
12,800 query--step groups and 122,768 original
query--rubric observations.
We classify each original rubric as unreachable,
informative, or saturated based on the binary
judgments of eight rollouts, and divide training
into five 40-step intervals.
For each group, we measure the occurrence of each
inactive state and its conditional rubric count,
averaged only over groups containing that state.

As shown in Figure~\ref{fig:inactivity-across-training}(a),
99.06\% of query groups contain at least one inactive
rubric, and 75.06\% contain \emph{both} unreachable
and saturated rubrics.
The co-occurrence rate remains high across all five
phases (73.52--76.41\%).
Unreachable-rubric occurrence remains near 84\%,
whereas saturated-rubric occurrence rises from
88.75\% to 91.48\%.
Figure~\ref{fig:inactivity-across-training}(b)
further shows that affected groups contain an average
of 2.88 unreachable or 3.79 saturated rubrics
over training.
These findings show that the two inactive states
are widespread and frequently coexist within the
same rollout group, motivating RuACT's state-dependent
use of rubrics as teachers and critics within
a shared training loop.

\subsection{Training Dynamics}
\label{app:training-dynamics}

\begin{figure}[t]
    \centering
    \includegraphics[width=\columnwidth]{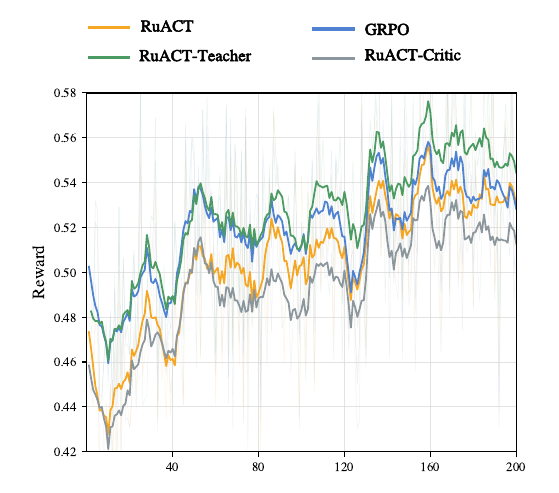}
    \caption{\textbf{Training reward on Qwen3-4B RaR-Medicine.}
    Reward magnitudes are not directly comparable
    across different scoring rules.}
    \label{fig:training-reward}
\end{figure}

\begin{figure}[t]
    \centering
    \includegraphics[width=\columnwidth]{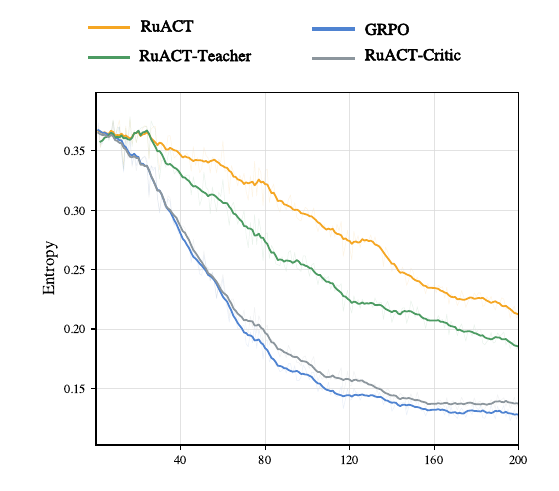}
    \caption{\textbf{Policy entropy during training.}
    Comparison over 200 training steps.}
    \label{fig:training-entropy}
\end{figure}

\begin{figure}[t]
    \centering
    \includegraphics[width=\columnwidth]{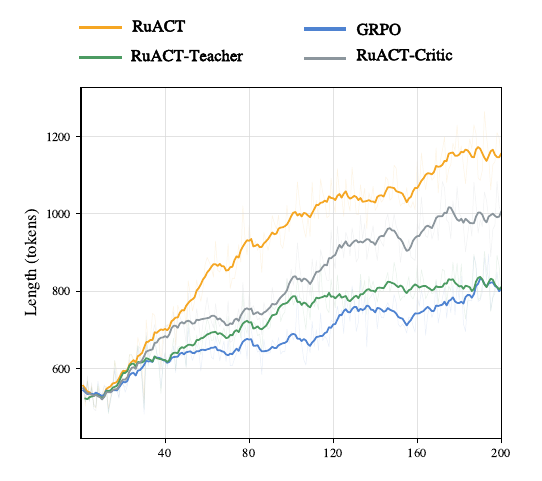}
    \caption{\textbf{Response length during training.}
    Comparison over 200 training steps.}
    \label{fig:training-response-length}
\end{figure}

\paragraph{RuACT Sustains Higher Policy Entropy.}
Figure~\ref{fig:training-reward}, \ref{fig:training-entropy}, \ref{fig:training-response-length} compares the training dynamics of GRPO, RuACT-Teacher, RuACT-Critic, and RuACT on Qwen3-4B RaR-Medicine.
While policy entropy declines across all methods, RuACT maintains substantially higher entropy throughout most of training.
RuACT-Teacher also exhibits slower entropy decay than GRPO, whereas RuACT-Critic follows a trajectory closer to standard GRPO.
Meanwhile, RuACT produces consistently longer responses as training progresses, suggesting more sustained behavioral exploration under recovered supervision.
However, greater entropy and response length do not necessarily imply higher response quality.
The reward curves also exhibit different magnitudes and fluctuations, but their absolute values are not directly comparable, since RuACT and RuACT-Critic incorporate refined child-rubric rewards while GRPO and RuACT-Teacher retain the original reward formulation.

\section{Policy-level Capability analysis}
\label{app:judge-robustness}

\subsection{Robustness to Evaluation Judges}
To assess sensitivity to the evaluation judge, we
re-evaluate the main-experiment models using Qwen3.7-Max.
As shown in Table~\ref{tab:judge-robustness},
RuACT retains the highest average score at both
model scales, raising GRPO from 44.16\% to 46.29\% on Qwen3-1.7B and from 55.95\% to 58.62\% on Qwen3-4B, and exceeding the strongest baseline, RuscaRL, at both scales (44.84\% and 56.31\%).
Although individual benchmark rankings vary
across judges, RuACT's overall advantage persists,
suggesting that its effectiveness is not specific
to a single evaluation judge.


\begin{table*}[t]

\centering
\begingroup

\fontsize{9.2}{11.2}\selectfont
\setlength{\tabcolsep}{3.5pt}
\renewcommand{\arraystretch}{1.0}
\setlength{\extrarowheight}{0pt}

\definecolor{HeaderGray}{HTML}{EFEFEF}
\definecolor{GroupGray}{HTML}{62666F}
\definecolor{OursBlue}{HTML}{285F91}
\definecolor{OursShade}{HTML}{F1F6FA}

\newcommand{\judgevanilla}{%
  {\fontsize{7.7}{9}\selectfont
  \textcolor{GroupGray}{\textbf{Vanilla}}}%
}

\newcommand{\judgebaselines}{%
  {\fontsize{7.7}{9}\selectfont
  \textcolor{GroupGray}{\textbf{Baselines}}}%
}

\newcommand{\judgeours}{%
  {\fontsize{7.7}{9}\selectfont
  \textcolor{OursBlue}{\textbf{Ours}}}%
}

\begin{tabularx}{\textwidth}{
>{\raggedright\arraybackslash}m{1.78cm}
!{\vrule width 0.4pt}
>{\centering\arraybackslash}m{1.22cm}
!{\vrule width 0.4pt}
>{\raggedright\arraybackslash}m{2.20cm}
!{\vrule width 0.4pt}
>{\centering\arraybackslash}X
>{\centering\arraybackslash}X
!{\vrule width 0.4pt}
>{\centering\arraybackslash}X
>{\centering\arraybackslash}X
!{\vrule width 0.4pt}
>{\centering\arraybackslash}X
}

\toprule[1.0pt]


\rowcolor{HeaderGray}
&
&
&
\multicolumn{2}
{c!{\vrule width 0.4pt}}
{\textbf{Medicine}}
&
\multicolumn{2}
{c!{\vrule width 0.4pt}}
{\textbf{Science}}
&
\\

\rowcolor{HeaderGray}
\multirow{-2}{*}{\textbf{Model}}
&
\multirow{-2}{*}{\textbf{Group}}
&
\multirow{-2}{*}{\textbf{Approach}}
&
\textbf{RaR-Med.}
&
\textbf{HealthBench}
&
\textbf{RaR-Sci.}
&
\textbf{ResearchQA}
&
\multirow{-2}{*}{\textbf{Avg.}}
\\

\midrule[0.7pt]



\multirow[c]{7}{*}{\textbf{Qwen3-1.7B}}
& \judgevanilla
& Base
& 41.34
& 20.88
& 49.92
& 48.13
& 40.07
\\

\cmidrule(lr){2-8}


& \multirow[c]{3}{*}{\judgebaselines}
& GRPO
& 46.08
& 24.80
& 53.06
& 52.71
& 44.16
\\

&
& Reweight
& 45.60
& 24.50
& 53.44
& 53.47
& 44.25
\\

&
& RuscaRL
& \underline{46.51}
& 21.29
& 53.13
& \underline{58.44}
& 44.84
\\

\cmidrule(lr){2-8}


& \multirow[c]{3}{*}{\judgeours}
& RuACT-Teacher
& 46.44
& 23.69
& 54.78
& 55.71
& 45.15
\\

&
& RuACT-Critic
& 45.48
& \underline{25.11}
& \textbf{55.23}
& \textbf{58.65}
& \underline{46.12}
\\

\rowcolor{OursShade}
&
& \textbf{RuACT}
& \textbf{46.62}
& \textbf{26.31}
& \underline{54.88}
& 57.35
& \textbf{46.29}
\\

\midrule[0.7pt]



\multirow[c]{7}{*}{\textbf{Qwen3-4B}}
& \judgevanilla
& Base
& 54.35
& 32.84
& 63.71
& 56.01
& 51.73
\\

\cmidrule(lr){2-8}


& \multirow[c]{3}{*}{\judgebaselines}
& GRPO
& 57.46
& 35.62
& 67.98
& 62.74
& 55.95
\\

&
& Reweight
& \underline{58.98}
& 35.81
& 66.87
& 63.55
& 56.30
\\

&
& RuscaRL
& 58.96
& 35.69
& 66.69
& 63.88
& 56.31
\\

\cmidrule(lr){2-8}


& \multirow[c]{3}{*}{\judgeours}
& RuACT-Teacher
& 58.71
& 37.72
& 67.36
& \underline{64.18}
& 56.99
\\

&
& RuACT-Critic
& 58.40
& \underline{37.91}
& \textbf{68.12}
& 63.80
& \underline{57.06}
\\

\rowcolor{OursShade}
&
& \textbf{RuACT}
& \textbf{60.04}
& \textbf{38.75}
& \underline{67.99}
& \textbf{67.68}
& \textbf{58.62}
\\

\bottomrule[1.0pt]

\end{tabularx}

\endgroup
\caption{\textbf{Robustness to evaluation judges.}
Results on medical and scientific reasoning benchmarks
re-evaluated using Qwen3.7-Max.
Avg. denotes the unweighted mean across four benchmarks.
Bold and underlined values indicate the best and
second-best results within each model scale.}
\label{tab:judge-robustness}
\end{table*}

\subsection{Policy-Level Capability Analysis}
\label{app:capability-analysis}

We provide additional details on the capability analysis
in Figure~\ref{fig:capability-map}.
The evaluation uses Qwen3-4B checkpoints on RaR-Medicine,
with $G=8$ unguided responses sampled per query for each
method. Both axes measure percentage-point improvements
relative to GRPO, using fixed evaluation inputs and
consistent rubric judgments.

\paragraph{Unreachable-Rate Reduction.}
The horizontal axis measures the reduction in unreachable
rubrics over the full test set, comprising 21,500
query--rubric pairs.
For method $m$, let $s^{(m)}_{xij}\in\{0,1\}$ denote
whether response $i$ satisfies original rubric $j$.
We compute
\begin{equation}
\begin{aligned}
U_m &= \frac{100}{|\mathcal T|}
\sum_{(x,j)\in\mathcal T}
\mathbb I\left[
\sum_{i=1}^{G}s^{(m)}_{xij}=0
\right],\\
X_m &= U_{\mathrm{GRPO}}-U_m,
\end{aligned}
\label{eq:app-unreachable-reduction}
\end{equation}
where $\mathcal T$ denotes all test query--rubric
pairs. Thus, this metric measures the overall reduction
in unreachable incidence, rather than recovery on a
fixed initially unreachable cohort.

\paragraph{Child-Rubric Satisfaction Gain.}
The vertical axis evaluates finer-grained satisfaction
on a fixed cohort $\mathcal S$ of 1,954 parent rubrics
identified as saturated under the Base model.
Child rubrics are generated using a fixed Qwen3.6
proposal model and shared across all evaluated methods.
Let $\mathcal C_{xj}$ denote the shared children of
parent $(x,j)$, and let $c^{(m)}_{xijk}$ be the binary
satisfaction judgment for child $k$.
We define
\begin{equation}
\begin{aligned}
C_m &=
\frac{100}{|\mathcal S|}
\sum_{(x,j)\in\mathcal S}
\frac{1}{G|\mathcal C_{xj}|}
\sum_{i=1}^{G}
\sum_{k\in\mathcal C_{xj}}
c^{(m)}_{xijk},\\
Y_m &= C_m-C_{\mathrm{GRPO}}.
\end{aligned}
\label{eq:app-child-satisfaction}
\end{equation}
Using the same parents and children ensures that
all methods are evaluated against identical
fine-grained requirements.
Unlike reward discrimination, this metric measures
the policy's satisfaction of these requirements,
rather than the variance of its training rewards.


\section{Case Studies}
\label{app:case-studies}

\subsection{Generation Quality Comparison across Methods}
\label{app:response-comparison}

We compare Base, GRPO, and RuACT on the same RaR-Medicine query, using the original ten rubrics rather than any child-rubric refinement. For each method, eight unguided rollouts are judged against the same requirements. The comparison table below reports per-rubric pass counts across the eight responses; we then present one recorded representative response per method.

\begingroup
\definecolor{CasePaper}{HTML}{F4F5F7}
\definecolor{CaseBase}{HTML}{FAE9EA}
\definecolor{CaseGRPO}{HTML}{EAF0FA}
\definecolor{CaseFull}{HTML}{E7F5EB}
\definecolor{CaseBad}{HTML}{A71929}
\definecolor{CaseGood}{HTML}{08793B}
\definecolor{CaseBaseLine}{HTML}{A94C53}
\definecolor{CaseGRPOLine}{HTML}{426FA7}
\definecolor{CaseFullLine}{HTML}{318456}
\newcommand{\casebad}[1]{\textcolor{CaseBad}{\textbf{\underline{#1}}}}
\newcommand{\casegood}[1]{\textcolor{CaseGood}{\textbf{\underline{#1}}}}

\begin{tcolorbox}[colback=CasePaper,colframe=black!20,boxrule=0.4pt,arc=1.2mm,
 left=2mm,right=2mm,top=1.4mm,bottom=1.4mm]
\textbf{Query.} A 39-year-old male presents with chronic knee pain, headache, nausea, epigastric pain, blurred vision, confusion, slowed speech, progressive decline in GCS leading to intubation, and seizure activity. CT brain reveals cerebral oedema, with an ammonia level of 652 mmol/L, greatly elevated urine orotic acid, high plasma glutamine level, and low arginine level. \textbf{Considering these findings, what is the pattern of inheritance for this disease?}
\end{tcolorbox}

\noindent\textbf{Original rubric-level comparisons.} Counts indicate satisfied responses out of eight per method (\emph{not} the three selected response scores).

\begingroup
\fontsize{7.6}{9.5}\selectfont
\setlength{\tabcolsep}{2.2pt}
\renewcommand{\arraystretch}{1.0}
\setlength{\extrarowheight}{0pt}
\noindent\begin{tabularx}{\columnwidth}{@{}>{\centering\arraybackslash}p{0.28cm}>{\raggedright\arraybackslash}X>{\centering\arraybackslash}p{0.65cm}>{\centering\arraybackslash}p{0.80cm}>{\centering\arraybackslash}p{0.87cm}@{}}
\toprule[0.8pt]
\rowcolor{CasePaper}\textbf{\#} & \textbf{Original rubric requirement} & \textbf{Base} & \textbf{GRPO} & \textbf{RuACT} \\
\midrule[0.5pt]
1 & \textbf{Identifies Inheritance Pattern.} Includes a clear statement identifying the pattern of inheritance as X-linked recessive. & \textcolor{CaseBad}{\textbf{0/8}} & 6/8 & \textcolor{CaseGood}{\textbf{8/8}} \\
\addlinespace[0.9pt]
2 & \textbf{Connects Symptoms to OTC Deficiency.} Explains how the symptoms, especially elevated ammonia and cerebral oedema, are indicative of ornithine transcarbamylase (OTC) deficiency. & \textcolor{CaseBad}{\textbf{0/8}} & 5/8 & \textcolor{CaseGood}{\textbf{8/8}} \\
\addlinespace[0.9pt]
3 & \textbf{Links Lab Findings to Disease.} Clearly links the lab findings such as high ammonia, elevated urine orotic acid, high plasma glutamine, and low arginine to urea cycle disorder. & \textcolor{CaseGood}{\textbf{8/8}} & \textcolor{CaseGood}{\textbf{8/8}} & \textcolor{CaseGood}{\textbf{8/8}} \\
\addlinespace[0.9pt]
4 & \textbf{Explains X-linked Recessive.} Provides an explanation of why and how OTC deficiency results from an X-linked recessive inheritance pattern. & \textcolor{CaseBad}{\textbf{0/8}} & 4/8 & \textcolor{CaseGood}{\textbf{8/8}} \\
\addlinespace[0.9pt]
5 & \textbf{Mentions Urea Cycle Disorder.} Identifies the disease as a urea cycle disorder, specifically OTC deficiency. & \textcolor{CaseBad}{\textbf{0/8}} & 5/8 & \textcolor{CaseGood}{\textbf{8/8}} \\
\addlinespace[0.9pt]
6 & \textbf{Describes Cerebral Oedema.} Describes the significance of cerebral oedema in the context of elevated ammonia levels. & 2/8 & 7/8 & \textcolor{CaseGood}{\textbf{8/8}} \\
\addlinespace[0.9pt]
7 & \textbf{Notes Arginine Role.} Mentions the role of low arginine levels in the symptoms or diagnosis of OTC deficiency. & 3/8 & 7/8 & \textcolor{CaseGood}{\textbf{8/8}} \\
\addlinespace[0.9pt]
8 & \textbf{Identifies Diagnostic Patterns.} Points out diagnostic patterns or key distinguishing factors of OTC deficiency among urea cycle disorders. & \textcolor{CaseBad}{\textbf{0/8}} & 6/8 & 6/8 \\
\addlinespace[0.9pt]
9 & \textbf{Clarifies Gender Susceptibility.} Clarifies how X-linked recessive patterns affect gender susceptibility, particularly noting male presentation. & 1/8 & 6/8 & \textcolor{CaseGood}{\textbf{8/8}} \\
\addlinespace[0.9pt]
10 & \textbf{Cautions Misidentification.} Does not mention the risk of misidentifying the inheritance pattern as autosomal. & 1/8 & 5/8 & 7/8 \\

\bottomrule[0.8pt]
\end{tabularx}
\endgroup
\endgroup
\vspace{4pt}

\begin{center}
\begingroup
\definecolor{CasePaper}{HTML}{F4F5F7}
\definecolor{CaseBase}{HTML}{FAE9EA}
\definecolor{CaseGRPO}{HTML}{EAF0FA}
\definecolor{CaseFull}{HTML}{E7F5EB}
\definecolor{CaseBad}{HTML}{A71929}
\definecolor{CaseGood}{HTML}{08793B}
\setlength{\tabcolsep}{4pt}
\fontsize{8.0}{9.5}\selectfont
\begin{tabularx}{\columnwidth}{@{}>{\raggedright\arraybackslash}Xcc@{}}
\toprule[0.8pt]
\rowcolor{CasePaper}\textbf{Selected response} & \textbf{Rubrics passed} & \textbf{Weighted score} \\
\midrule[0.5pt]
\rowcolor{CaseBase}Base (sample 0) & 1/10 & 0.11 \\
\rowcolor{CaseGRPO}GRPO (sample 5) & 6/10 & 0.53 \\
\rowcolor{CaseFull}\textbf{RuACT} (sample 0) & 10/10 & 1.00 \\
\bottomrule[0.8pt]
\end{tabularx}
\endgroup
\end{center}

\paragraph{Representative recorded responses.}
The responses below reproduce the selected generated texts in full, with paragraph and list formatting normalized. Red emphasis marks an incorrect diagnosis or inheritance claim; green emphasis marks correct rubric-relevant evidence. An emphasized passage does not imply that all remaining medical details have been independently verified.

\begingroup
\definecolor{CaseBase}{HTML}{FAE9EA}
\definecolor{CaseGRPO}{HTML}{EAF0FA}
\definecolor{CaseFull}{HTML}{E7F5EB}
\definecolor{CaseBad}{HTML}{A71929}
\definecolor{CaseGood}{HTML}{08793B}
\definecolor{CaseBaseLine}{HTML}{A94C53}
\definecolor{CaseGRPOLine}{HTML}{426FA7}
\definecolor{CaseFullLine}{HTML}{318456}
\newcommand{\casebad}[1]{\textcolor{CaseBad}{\textbf{\underline{#1}}}}
\newcommand{\casegood}[1]{\textcolor{CaseGood}{\textbf{\underline{#1}}}}
\providecommand{\tightlist}{\setlength{\itemsep}{0pt}\setlength{\parskip}{0pt}}
\setlength{\emergencystretch}{2em}
\begin{tcolorbox}[enhanced,breakable,colback=white,colframe=CaseBaseLine,boxrule=0.55pt,
 colbacktitle=CaseBase,coltitle=black,fonttitle=\bfseries,arc=0.8mm,
 title={Base \hfill Sample 0: 1/10 rubrics},
 left=2.2mm,right=2.2mm,top=1.7mm,bottom=1.7mm,before skip=6pt,after skip=7pt]
\fontsize{8.2}{10.1}\selectfont
The clinical and laboratory findings in this 39-year-old male are highly suggestive of \casebad{argininemia}, a rare autosomal \textbf{recessive} metabolic disorder.

\textbf{Key Features of Argininemia:}

\begin{itemize}
\tightlist
\item
  \textbf{Clinical Presentation}:

  \begin{itemize}
  \tightlist
  \item
    Chronic kidney disease (due to urea cycle defect)
  \item
    Neurological symptoms (headache, nausea, epigastric pain, blurred vision, confusion, slowed speech, decreased GCS, seizures)
  \item
    Coma and intubation
  \end{itemize}
\item
  \textbf{Laboratory Findings}:

  \begin{itemize}
  \tightlist
  \item
    \textbf{Elevated ammonia levels} (652 mmol/L is extremely high, indicating severe hyperammonemia)
  \item
    \textbf{Elevated urine orotic acid}
  \item
    \textbf{High plasma glutamine} (a hallmark of urea cycle disorders)
  \item
    \textbf{Low plasma arginine} (due to impaired conversion of glutamine to arginine)
  \end{itemize}
\item
  \textbf{Genetic Basis}:

  \begin{itemize}
  \tightlist
  \item
    Argininemia is caused by \textbf{deficiency of the enzyme arginase}, which is located on \textbf{chromosome 12}.
  \item
    \textbf{Arginase deficiency} is an \casebad{autosomal recessive} disorder, meaning it is inherited when both parents are carriers of the defective gene.
  \end{itemize}
\end{itemize}

\textbf{Summary:}

\begin{itemize}
\tightlist
\item
  \textbf{Pattern of inheritance}: \casebad{Autosomal recessive}
\item
  \textbf{Cause}: Deficiency of \textbf{arginase} (a key enzyme in the urea cycle)
\item
  \textbf{Clinical course}: Progressive neurological decline, hyperammonemia, and metabolic encephalopathy
\end{itemize}

\textbf{Answer:}

\casebad{Autosomal recessive}.

\end{tcolorbox}

\begin{tcolorbox}[enhanced,breakable,colback=white,colframe=CaseGRPOLine,boxrule=0.55pt,
 colbacktitle=CaseGRPO,coltitle=black,fonttitle=\bfseries,arc=0.8mm,
 title={GRPO \hfill Sample 5: 6/10 rubrics},
 left=2.2mm,right=2.2mm,top=1.7mm,bottom=1.7mm,before skip=6pt,after skip=7pt]
\fontsize{8.2}{10.1}\selectfont
The clinical and laboratory findings in this 39-year-old male are highly suggestive of \textbf{argininemia} (also known as \textbf{argininosuccinic aciduria} or \textbf{arginine metabolism disorder}), which is a rare inherited metabolic disorder.

\textbf{Let's break down the findings:}

\textbf{Clinical Presentation:}

\begin{itemize}
\tightlist
\item
  \textbf{Chronic knee pain}: Possible due to \textbf{hyperammonemia} leading to \textbf{urea cycle disorder} symptoms (e.g., muscle weakness, joint pain).
\item
  \textbf{Headache, nausea, epigastric pain, blurred vision, confusion, slowed speech}: Classic symptoms of \textbf{hyperammonemia} (elevated blood ammonia levels), which can cause \textbf{central nervous system (CNS) toxicity}.
\item
  \textbf{Progressive decline in GCS leading to intubation and seizure activity}: Severe \textbf{hyperammonemia} can lead to \textbf{CNS depression}, \textbf{encephalopathy}, and \textbf{seizures}.
\item
  \textbf{Seizure activity}: Further supports \textbf{hyperammonemia} and \textbf{CNS involvement}.
\end{itemize}

\textbf{Imaging:}

\begin{itemize}
\tightlist
\item
  \textbf{CT brain reveals cerebral edema}: Consistent with \textbf{hyperammonemia} (ammonia toxicity to the brain).
\end{itemize}

\textbf{Lab Findings:}

\begin{itemize}
\tightlist
\item
  \textbf{Ammonia level of 652 mmol/L}: \textbf{Severely elevated}, indicating \textbf{hyperammonemia}.
\item
  \textbf{Elevated urine orotic acid}: Orotic acid is a hallmark of \textbf{ornithine transcarbamylase deficiency} (a type of \textbf{urea cycle disorder}), but also seen in \textbf{argininemia}.
\item
  \textbf{High plasma glutamine level}: Glutamine is a byproduct of \textbf{ammonia metabolism} in \textbf{urea cycle disorders}.
\item
  \textbf{Low arginine level}: Arginine is a key substrate in the \textbf{urea cycle}, and its deficiency can occur in \textbf{argininemia} or \textbf{ornithine transcarbamylase deficiency}.
\end{itemize}

\textbf{Genetic Considerations:}

\textbf{Argininemia (Argininosuccinic aciduria):}

\begin{itemize}
\tightlist
\item
  \textbf{Cause}: Deficiency of \textbf{argininase} (enymes in the urea cycle).
\item
  \textbf{Inheritance}: \casebad{Autosomal recessive}.
\item
  \textbf{Pathophysiology}: Impaired breakdown of arginine leads to \textbf{accumulation of arginine} and \textbf{orotic acid}. Elevated ammonia results from \textbf{impaired urea cycle}.
\item
  \textbf{Clinical features}: \textbf{Hyperammonemia}, \textbf{encephalopathy}, \textbf{seizures}, \textbf{knee pain} (due to \textbf{urea cycle dysfunction}).
\item
  \textbf{Urine orotic acid}: Yes, \textbf{elevated}.
\item
  \textbf{Plasma glutamine}: Yes, \textbf{elevated} (due to \textbf{ammonia toxicity} and \textbf{glutamine synthesis}).
\item
  \textbf{Low arginine}: Yes, due to \textbf{accumulation} of arginine in the blood (but \textbf{arginase} deficiency leads to \textbf{accumulation} of \textbf{arginine} and \textbf{orotic acid}).
\end{itemize}

\textbf{Ornithine Transcarbamylase Deficiency (OTCD):}

\begin{itemize}
\tightlist
\item
  \textbf{Cause}: Deficiency of \textbf{ornithine transcarbamylase} (enzyme in the urea cycle).
\item
  \textbf{Inheritance}: \casebad{Autosomal recessive}.
\item
  \textbf{Pathophysiology}: Impaired conversion of ornithine to carbamoyl phosphate, leading to \textbf{accumulation of ammonia}, \textbf{orotic acid}, \textbf{glutamine}, and \textbf{low arginine}.
\item
  \textbf{Urine orotic acid}: Yes, \textbf{elevated}.
\item
  \textbf{Plasma glutamine}: Yes, \textbf{elevated}.
\item
  \textbf{Low arginine}: Yes, due to \textbf{impaired urea cycle}.
\end{itemize}

\textbf{Key Distinguishing Features:}

\begin{itemize}
\tightlist
\item
  \textbf{Argininemia} is caused by \textbf{arginase deficiency}, leading to \textbf{accumulation of arginine} and \textbf{orotic acid}.
\item
  \casegood{OTCD} is caused by \textbf{ornithine transcarbamylase deficiency}, leading to \textbf{accumulation of ammonia}, \textbf{orotic acid}, \textbf{glutamine}, and \textbf{low arginine}.
\end{itemize}

\textbf{In this case:}

\begin{itemize}
\tightlist
\item
  \textbf{Elevated ammonia}, \textbf{elevated urine orotic acid}, \textbf{elevated glutamine}, \textbf{low arginine} --- \textbf{all consistent with either argininemia or OTCD}.
\item
  However, \textbf{arginine level} is \casegood{low}, which is more consistent with \casegood{OTCD} (since arginine is a substrate in the urea cycle, and its deficiency is due to \textbf{impaired cycle}), whereas in \textbf{argininemia}, \textbf{arginine is elevated} (but in this case, it's \casegood{low}, so this suggests \casegood{OTCD}).
\end{itemize}

Wait --- this is a critical point. Let's clarify:

\begin{itemize}
\tightlist
\item
  \textbf{Argininemia}: \textbf{Arginine is elevated} (because arginase breaks down arginine, and if it's deficient, arginine accumulates).
\item
  \casegood{OTCD}: \textbf{Arginine is low} (because the enzyme that converts ornithine to carbamoyl phosphate is deficient, leading to \textbf{accumulation of ammonia}, and \textbf{arginine is a product of the urea cycle}, so its levels are \casegood{low}).
\end{itemize}

In this patient, \textbf{arginine is low}, so this \textbf{points towards OTCD}, not argininemia.

But wait --- the patient has \textbf{elevated urine orotic acid}. Orotic acid is a \textbf{hallmark of OTCD} (due to \textbf{deficiency of ornithine transcarbamylase}, which leads to \textbf{orotic aciduria}).

So, the most likely diagnosis is \casegood{Ornithine Transcarbamylase Deficiency (OTCD)}, which is \casebad{autosomal recessive}.

However, \textbf{argininemia} can also present with \textbf{elevated orotic acid} (due to \textbf{deficiency of arginase}, which leads to \textbf{accumulation of orotic acid}).

But in this case, \textbf{arginine is low}, which is \textbf{more consistent with OTCD}.

\textbf{Conclusion:}

The \textbf{pattern of inheritance} for this disease is \casebad{autosomal recessive}.

\textbf{Final Answer:}

\casebad{Autosomal recessive} (specifically, \textbf{Ornithine Transcarbamylase Deficiency} or \textbf{Argininemia}, but given the low arginine and elevated orotic acid, \casegood{OTCD} is more likely).

\textbf{Answer: Autosomal recessive}.

\end{tcolorbox}

\begin{tcolorbox}[enhanced,breakable,colback=white,colframe=CaseFullLine,boxrule=0.55pt,
 colbacktitle=CaseFull,coltitle=black,fonttitle=\bfseries,arc=0.8mm,
 title={RuACT \hfill Sample 0: 10/10 rubrics},
 left=2.2mm,right=2.2mm,top=1.7mm,bottom=1.7mm,before skip=6pt,after skip=7pt]
\fontsize{8.2}{10.1}\selectfont
The case presents a 39-year-old male with chronic knee pain, headache, nausea, epigastric pain, blurred vision, confusion, slowed speech, progressive decline in GCS leading to intubation, and seizure activity. The CT brain shows cerebral edema, with elevated ammonia levels (652 mmol/L), increased urine orotic acid, high plasma glutamine levels, and low arginine levels. These findings are highly suggestive of \textbf{urea cycle disorder (UCD)}, specifically \textbf{argininemia} or \textbf{ornithine transcarbamylase deficiency}. Let's break down the diagnosis and inheritance pattern.

\textbf{Key Findings and Diagnosis:}

\begin{enumerate}
\def\labelenumi{\arabic{enumi}.}
\tightlist
\item
  \textbf{Elevated Ammonia (652 mmol/L):}

  \begin{itemize}
  \tightlist
  \item
    Ammonia toxicity is a hallmark of urea cycle disorders, where the body cannot effectively remove ammonia from the blood, leading to its accumulation.
  \item
    High ammonia levels can cause cerebral edema, encephalopathy, seizures, and coma.
  \end{itemize}
\item
  \textbf{Elevated Urine Orotic Acid:}

  \begin{itemize}
  \tightlist
  \item
    Orotic acid is a substrate in the urea cycle. Its elevation is classically seen in \casegood{ornithine transcarbamylase (OTC) deficiency}, which is the most common urea cycle disorder.
  \item
    Orotic acid is also elevated in \textbf{argininemia} (due to defects in the argininosuccinate lyase or argininosuccinate synthase enzymes).
  \end{itemize}
\item
  \textbf{High Plasma Glutamine:}

  \begin{itemize}
  \tightlist
  \item
    Glutamine accumulates in UCDs because the urea cycle cannot process nitrogen, leading to its conversion to glutamine via the glutaminase pathway.
  \item
    This is seen in \textbf{argininemia}, \textbf{citrullinemia}, and \casegood{OTC deficiency}.
  \end{itemize}
\item
  \textbf{Low Arginine Levels:}

  \begin{itemize}
  \tightlist
  \item
    Arginine is a product of the urea cycle. Low levels are seen in \casegood{OTC deficiency} (since OTC is deficient and cannot convert carbamyl phosphate to arginine) and \textbf{argininemia} (due to accumulation of downstream metabolites).
  \end{itemize}
\item
  \textbf{Seizures, Encephalopathy, and Cerebral Edema:}

  \begin{itemize}
  \tightlist
  \item
    These are classic manifestations of ammonia toxicity in UCDs, particularly in \casegood{OTC deficiency} and \textbf{argininemia}.
  \end{itemize}
\item
  \textbf{Chronic Knee Pain:}

  \begin{itemize}
  \tightlist
  \item
    This could be due to \textbf{hyperammonemia} leading to metabolic disturbances, or it could be a secondary manifestation of \textbf{urea cycle disorder} (e.g., in \textbf{argininemia}, there is also a defect in the argininosuccinate lyase enzyme, which can lead to metabolic acidosis and other systemic issues).
  \end{itemize}
\end{enumerate}

\textbf{Differential Diagnosis:}

\begin{itemize}
\tightlist
\item
  \textbf{Ornithine Transcarbamylase Deficiency (OTC Deficiency):}

  \begin{itemize}
  \tightlist
  \item
    Most common urea cycle disorder.\\
  \item
    Deficiency in OTC, which is critical for the final step of the urea cycle (conversion of carbamyl phosphate to arginine).\\
  \item
    Leads to accumulation of ammonia, ornithine, and other intermediates.\\
  \item
    Low arginine, high glutamine, and elevated orotic acid.\\
  \item
    X-linked inheritance (due to location on the X chromosome).
  \end{itemize}
\item
  \textbf{Argininemia (Argininosuccinate Lyase Deficiency):}

  \begin{itemize}
  \tightlist
  \item
    Deficiency in argininosuccinate lyase, leading to accumulation of argininosuccinate and citrulline.\\
  \item
    Can also cause elevated ammonia and glutamine.\\
  \item
    Autosomal recessive inheritance.
  \end{itemize}
\item
  \textbf{Other Urea Cycle Disorders:}

  \begin{itemize}
  \tightlist
  \item
    \textbf{Citrullinemia} (argininosuccinate synthase deficiency), \textbf{hyperammonemia with impaired urea cycle}, etc.
  \end{itemize}
\end{itemize}

\textbf{Inheritance Pattern:}

\begin{itemize}
\tightlist
\item
  \textbf{Ornithine Transcarbamylase Deficiency (OTC Deficiency):}

  \begin{itemize}
  \tightlist
  \item
    \casegood{X-linked recessive inheritance.}\\
  \item
    Since the gene is located on the X chromosome, males (who have only one X chromosome) are more frequently affected.\\
  \item
    Females can be carriers but may have milder symptoms or be asymptomatic.
  \end{itemize}
\item
  \textbf{Argininemia (Argininosuccinate Lyase Deficiency):}

  \begin{itemize}
  \tightlist
  \item
    \textbf{Autosomal recessive inheritance.}\\
  \item
    Requires two copies of the defective gene (one from each parent).\\
  \item
    Both males and females are equally affected.
  \end{itemize}
\end{itemize}

\textbf{Clinical Course and Management:}

\begin{itemize}
\tightlist
\item
  \textbf{Acute presentation:}

  \begin{itemize}
  \tightlist
  \item
    Ammonia toxicity leads to encephalopathy, seizures, and cerebral edema.\\
  \item
    Immediate management includes \textbf{lavage of the gastrointestinal tract}, \textbf{administration of lactulose or protein restriction}, \textbf{arginine supplementation}, and \textbf{hemodialysis} if necessary.
  \end{itemize}
\item
  \textbf{Long-term management:}

  \begin{itemize}
  \tightlist
  \item
    \textbf{Low-protein diet} to reduce nitrogen load.\\
  \item
    \textbf{Supplementation with arginine} and \textbf{orotic acid}.\\
  \item
    \textbf{Regular monitoring of ammonia levels} and \textbf{genetic counseling}.
  \end{itemize}
\item
  \textbf{Prognosis:}

  \begin{itemize}
  \tightlist
  \item
    Without treatment, \casegood{OTC deficiency} can lead to \textbf{coma, brain damage, or death}.\\
  \item
    \textbf{Argininemia} can also be severe, but with early diagnosis and management, outcomes can be improved.
  \end{itemize}
\end{itemize}

\textbf{Conclusion:}

The patient's presentation is consistent with \textbf{ornithine transcarbamylase deficiency (OTC deficiency)}, an \casegood{X-linked recessive} disorder. The key features---\textbf{elevated ammonia, urine orotic acid, high glutamine, low arginine}, and \textbf{cerebral edema}---are classic for this condition. The \casegood{X-linked inheritance} pattern is due to the gene's location on the X chromosome, making males more susceptible.

\textbf{Answer:} The disease follows an \casegood{X-linked recessive} inheritance pattern.

\end{tcolorbox}

\endgroup

\noindent\textbf{Interpretation.}
The selected Base response identifies argininemia and gives an autosomal-recessive answer (1/10 rubrics). GRPO recognizes multiple biochemical clues pointing to OTC deficiency, but its final inheritance classification remains autosomal recessive (6/10). RuACT makes the X-linked recessive determination and links it to the relevant biochemical pattern (10/10). Across all eight rollouts per method, the mean binary satisfaction rates are 18.75\%, 73.75\%, and 96.25\% for Base, GRPO, and RuACT, respectively. The case illustrates differences in rubric satisfaction, not independent expert validation of every medical statement.

\subsection{Child-Rubric Refinement under Saturation}
\label{app:critic-case}

A parent rubric concerning abdominal-muscle involvement is satisfied
by every sampled response, yet finer requirements about individual
muscles yield non-uniform binary judgments. Table~\ref{tab:critic-case}
shows the logged judgments alongside excerpts restricted to the
rubric-relevant portions of all eight responses.

\begin{table*}[t]
\centering
\begingroup
\definecolor{RCHeader}{HTML}{EEEEF1}
\definecolor{RCParent}{HTML}{FAF1F0}
\definecolor{RCBlue}{HTML}{285F91}
\definecolor{RCGray}{HTML}{888888}
\fontsize{8.8}{10.5}\selectfont
\setlength{\tabcolsep}{4pt}
\renewcommand{\arraystretch}{1.0}
\setlength{\extrarowheight}{0pt}

\begin{minipage}{\textwidth}
\textbf{Query.} Which muscles are specifically involved in the process of expiration?\par\vspace{4pt}
\textbf{Saturated parent (P; 8/8).} Explains that abdominal muscles contract to increase intra-abdominal pressure during active expiration.\par\vspace{5pt}
\textbf{Refined child rubrics.}\quad
\textcolor{RCBlue}{\textbf{$c_1$}}: External oblique contraction increases abdominal pressure during forced exhalation;\quad
\textcolor{RCBlue}{\textbf{$c_2$}}: Internal oblique helps compress the abdomen during active expiration;\quad
\textcolor{RCBlue}{\textbf{$c_3$}}: Transversus abdominis compresses abdominal contents to raise intra-abdominal pressure.\par
\end{minipage}
\vspace{7pt}

\newcommand{\rcpass}{\textcolor{RCBlue}{\textbf{1}}}
\newcommand{\rcfail}{\textcolor{RCGray}{0}}
\begin{tabularx}{\textwidth}{@{}
>{\centering\arraybackslash}p{0.62cm}
>{\raggedright\arraybackslash}X
>{\centering\arraybackslash}p{0.61cm}
>{\centering\arraybackslash}p{0.61cm}
>{\centering\arraybackslash}p{0.61cm}
>{\centering\arraybackslash}p{0.61cm}
@{}}
\toprule[0.9pt]
\rowcolor{RCHeader}
\textbf{ID} & \textbf{Rubric-relevant response excerpts} & \textbf{P} & \textbf{$c_1$} & \textbf{$c_2$} & \textbf{$c_3$} \\
\midrule[0.65pt]

R1 & ``\textbf{External Oblique}: Contracts to pull the ribcage downward and inward (indirectly aiding expiration). \ldots{} \textbf{Internal Oblique}: Also helps in pulling the ribcage inward. \ldots{} \textbf{Transverse Abdominis}: Contracts to increase abdominal pressure, which helps push the diaphragm upward.''
& \rcpass & \rcfail & \rcpass & \rcpass \\
\midrule[0.25pt]

R2 & ``\textbf{Transverse Abdominis}: Contracts to compress the abdominal cavity, helping push the diaphragm upward. \ldots{} \textbf{Oblique Muscles} (Internal and External Obliques): Contract to stabilize the abdominal wall and assist in forced expiration.''
& \rcpass & \rcfail & \rcfail & \rcfail \\
\midrule[0.25pt]

R3 & ``\textbf{External and Internal Obliques}: Contract to compress the abdominal cavity, further helping to expel air from the lungs. \ldots{} \textbf{Transverse Abdominis}: Contracts to increase intra-abdominal pressure, assisting in expiration.''
& \rcpass & \rcpass & \rcpass & \rcpass \\
\midrule[0.25pt]

R4 & ``\textbf{External Oblique}: Contracts to pull the ribcage inward and compress the abdominal cavity. \ldots{} \textbf{Internal Oblique}: Similarly contributes to abdominal compression. \ldots{} \textbf{Transversus Abdominis}: Contracts to increase abdominal pressure, helping push the diaphragm up.''
& \rcpass & \rcfail & \rcpass & \rcpass \\
\midrule[0.25pt]

R5 & ``\textbf{External Oblique} and \textbf{Internal Oblique}: Contract to compress the abdominal cavity, helping push the diaphragm upward. \ldots{} \textbf{Transversus Abdominis}: Contracts to increase abdominal pressure, assisting in expelling air.''
& \rcpass & \rcfail & \rcfail & \rcfail \\
\midrule[0.25pt]

R6 & ``\textbf{External and Internal Oblique Muscles}: Contract to compress the abdominal cavity, further increasing intra-abdominal pressure and aiding in expiration. \ldots{} \textbf{Transversus Abdominis}: Contracts to tighten the abdominal wall, assisting in expelling air from the lungs.''
& \rcpass & \rcpass & \rcpass & \rcpass \\
\midrule[0.25pt]

R7 & ``\textbf{External Oblique} and \textbf{Internal Oblique}: Contract to compress the abdominal cavity, helping push the diaphragm upward. \ldots{} \textbf{Transversus Abdominis}: Contracts to increase abdominal pressure, assisting in expelling air.''
& \rcpass & \rcfail & \rcfail & \rcfail \\
\midrule[0.25pt]

R8 & ``\textbf{Rectus Abdominis}: Contracts to push the abdominal organs inward, increasing intra-abdominal pressure and helping expel air from the lungs. \ldots{} \textbf{Transverse Abdominis} and \textbf{Oblique Abdominals} (external and internal): These muscles compress the abdominal cavity, further increasing intra-abdominal pressure and assisting in expiration.''
& \rcpass & \rcfail & \rcfail & \rcfail \\
\midrule[0.65pt]
\rowcolor{RCParent}
\textbf{Pass} & \textbf{Number of satisfying responses (out of eight)} & \textbf{8/8} & \textbf{2/8} & \textbf{4/8} & \textbf{4/8} \\
\bottomrule[0.9pt]
\end{tabularx}
\vspace{2pt}
{\footnotesize\color{RCGray}Numbers reproduce the stored rubric-judge verdicts (1 = satisfied; 0 = not satisfied). Excerpts retain original response wording but omit unrelated material, marked by ellipses.}\par
\endgroup
\caption{\textbf{Critic-side case study on RaR-Medicine.} A saturated parent is satisfied by all eight responses, whereas its finer child rubrics yield non-uniform judgments on the same rollout group.}
\label{tab:critic-case}
\end{table*}

\definecolor{RuACTPromptBlue}{HTML}{354C8C}
\definecolor{RuACTPromptGreen}{HTML}{27664F}
\definecolor{RuACTPromptOrange}{HTML}{946025}
\definecolor{RuACTPromptBackground}{HTML}{F8F9FB}

\newtcblisting{ruactprompt}[2]{%
  enhanced,
  breakable,
  listing only,
  listing engine=listings,
  colback=RuACTPromptBackground,
  colframe=#2,
  colbacktitle=#2,
  coltitle=white,
  title={#1},
  fonttitle=\bfseries\small,
  boxrule=0.55pt,
  arc=1.5mm,
  left=2.0mm,
  right=2.0mm,
  top=1.5mm,
  bottom=1.5mm,
  before skip=5pt,
  after skip=7pt,
  listing options={%
    basicstyle=\ttfamily\fontsize{7.5}{9}\selectfont,
    columns=fullflexible,
    breaklines=true,
    breakatwhitespace=false,
    keepspaces=true,
    showstringspaces=false,
    upquote=true,
    tabsize=2
  }
}

\section{Prompt Templates}
\label{app:prompts}

We provide the prompt templates used for rubric-conditioned teacher guidance, child-rubric proposal, and binary rubric evaluation. Runtime inputs are denoted by braced placeholders. The templates cover the positive-only and mixed-polarity teacher instructions, the initial and feedback-guided child-rubric proposal rounds, and the shared binary judgment format.

\subsection{Privileged Teacher Prompt}
\label{app:teacher-prompt}

For currently unreachable rubrics of either polarity, the teacher receives the original query and privileged feedback specifying up to three unmet requirements or penalized behaviors, ranked by absolute rubric weight. Positive-weight rubrics are expressed as \texttt{[REQUIRE]} instructions, and negative-weight rubrics as \texttt{[AVOID]} instructions. The rubric-guided response is used only for the overall-reward acceptance gate. For accepted groups, teacher--student forward KL is evaluated on identical prefixes of the highest-reward ordinary rollout, rather than by directly imitating the guided response (Section~\ref{sec:teacher}).

\begin{ruactprompt}{Teacher: Positive Requirements}{RuACTPromptGreen}
{original user prompt}

Private rubric feedback: every sampled answer failed the positive requirements below. Satisfy them while preserving correct content. Do not mention this feedback in the answer.
1. [REQUIRE] {failed positive rubric description}
2. [REQUIRE] {another failed positive rubric description}
...
\end{ruactprompt}

\paragraph{Polarity-aware guidance.}
The prompt is chosen according to the selected unreachable rubrics. When all selected rubrics have positive weights, the positive-only template above is used. When at least one negative-weight rubric is selected, the following mixed-polarity template is used instead. Only the selected \texttt{[REQUIRE]} and \texttt{[AVOID]} instructions are included.

\begin{ruactprompt}{Teacher: Requirements and Pitfalls}{RuACTPromptGreen}
{original user prompt}

Private rubric feedback: every sampled answer failed the required points below or committed the listed pitfalls. Satisfy every [REQUIRE] point and do not commit any [AVOID] pitfall, while preserving correct content. Do not mention this feedback in the answer.
1. [REQUIRE] {failed positive rubric description}
2. [AVOID] {failed negative-weight pitfall description}
...
\end{ruactprompt}

\subsection{Child Rubric Generation and Regeneration}
\label{app:critic-prompt}

For a saturated parent rubric, child candidates are obtained from a preverified library when available and supplemented by a proposal model conditioned on the query, parent title and description, and fixed few-shot demonstrations. The initial proposal does not observe current rollout responses or their judgments. Candidates must remain within the parent's objective and be independently judgeable against a single answer. If too few candidates pass judgment-based selection, at most one additional proposal round is performed using feedback on rejected candidates, without revealing any rollout response text. The demonstrations below are included through the corresponding runtime placeholder.

\begin{ruactprompt}{Child Proposal: Fixed Few-Shot Demonstrations}{RuACTPromptOrange}
Few-shot demonstrations (format and specificity only; do not copy their medical content):

Example 1
Query: A patient has acute chest pain. What is the initial evaluation?
Parent rubric: Identifies time-sensitive evaluation for life-threatening cardiac causes.
Valid output:
{"candidates":[
  {"title":"Serial ischemia testing","completion_point":"States that serial ECG and troponin assessment are needed when acute coronary syndrome remains a concern."},
  {"title":"Immediate instability check","completion_point":"Checks for hemodynamic instability before proceeding with routine evaluation."}
]}

Example 2
Query: How should suspected bacterial meningitis be managed initially?
Parent rubric: Recommends prompt empiric treatment for suspected bacterial meningitis.
Valid output:
{"candidates":[
  {"title":"Do not delay treatment","completion_point":"States that empiric antibiotics should not be delayed when lumbar puncture or imaging would postpone treatment."},
  {"title":"Blood cultures first when feasible","completion_point":"Requests blood cultures before antibiotics when this can be done without delaying treatment."}
]}

Example 3
Query: What discharge advice is appropriate after a new medication is started?
Parent rubric: Provides safe, actionable follow-up advice.
Valid output:
{"candidates":[
  {"title":"Specific warning signs","completion_point":"Names concrete symptoms that require urgent reassessment rather than giving only generic follow-up advice."},
  {"title":"Defined follow-up timing","completion_point":"Specifies when and with whom the patient should follow up."}
]}
\end{ruactprompt}

\begin{ruactprompt}{Child Proposal: Target Prompt}{RuACTPromptOrange}
You are refining one rubric for a RaR training example.

{few-shot demonstrations above}

Now solve the target example. Follow the demonstrated JSON shape, but derive every completion point only from the target query and parent rubric.

User query:
{query}

Parent rubric:
Title: {parent title}
Description: {parent description}
{polarity-specific instruction, if applicable}
{feedback on rejected child candidates, only in an additional proposal round}
Generate between {min_candidate_count} and {max_candidate_count} atomic completion-point candidates. Do not request, use, or describe any rollout response texts or per-response judgments. If aggregate feedback on rejected candidates is supplied, use it only to propose different, parent-faithful completion points.

For every candidate:
- Stay inside the exact objective and direction of the parent rubric.
- Express exactly one objective, factually defensible, query-relevant detail that a stronger answer should include to complete the parent requirement.
- The point may operationalize or add detail to the parent, but must not introduce a different evaluation objective.
- Prefer distinct refinement operators when appropriate, such as mechanism, evidence, boundary condition, consequence, management implication, operationalization, or a concrete example.
- Keep the requirement atomic and binary-judgeable from the query and one answer.
- Do not add vague demands such as being more comprehensive, detailed, clear, or accurate.
- Do not introduce unrelated topics, style preferences, response IDs, weights, polarity labels, scoring rules, or references to the parent.
- Make the candidates semantically distinct, not paraphrases of one another or of candidates identified as rejected in the additional-round feedback.
- Match the language of the parent rubric.

Return only a short standalone title and the single completion_point for each candidate.
\end{ruactprompt}

The proposal model outputs a JSON \texttt{candidates} array whose entries contain \texttt{title} and \texttt{completion\_point}. In the main configuration, up to five initial candidates are considered per parent. Candidate children are judged on the current rollout group and filtered for reward discrimination. When the number of eligible children is below the required minimum, an additional proposal round (at most one) is triggered. The second round uses the same query and parent requirement, augmented with feedback identifying rejected child proposals and their aggregate judgment outcomes; no rollout responses are shown to the proposal model. New candidates are judged and filtered using the same criterion. Up to three eligible children are retained per parent, subject to a maximum of three active children per query. If no eligible child is retained, the original parent contribution is preserved.

\subsection{Rubric Evaluation Prompt}
\label{app:judge-prompt}

The binary rubric judge evaluates a single public final answer against the query and numbered rubric descriptions, without observing reference answers, other responses, rubric weights, hidden reasoning, or training statistics. When applicable, the public answer is extracted after \texttt{</think>}. If no usable final answer can be extracted, all rubric verdicts are recorded as \texttt{NOT\_PRESENT}. Rubric weights are applied only during downstream reward aggregation.

\begin{ruactprompt}{Binary Rubric Judge: Shared Instruction Template}{RuACTPromptBlue}
Given a user prompt, a generated response, and a list of quality rubrics, evaluate the response against EACH rubric.
For each rubric:
- Mark "PRESENT" if the criterion is satisfied, or "NOT_PRESENT" if it is not. For example, given the response "Apples are red", the rubric "Mentions apples" is PRESENT, "Does not mention strawberries" is also PRESENT since the response does not mention strawberries, and "Mentions oranges" is NOT_PRESENT. Likewise, "Avoids mentioning strawberries" is PRESENT, while "Avoids mentioning apples" is NOT_PRESENT.
- If a rubric contains multiple sentences or criteria, consider all of them. If any required part is not met, mark it NOT_PRESENT. Return PRESENT only if the entire rubric is met.
- When a rubric gives examples with wording such as "such as", "for example", or "including", the response does not need to include every example. Satisfying one valid listed example can satisfy the criterion. There is no partial credit.
Return only one strict JSON object. Its keys must be the rubric numbers and its values must be exactly "PRESENT" or "NOT_PRESENT". Do not return Markdown, reasons, or extra keys.

The output object must contain exactly one key for each numbered rubric, with no omissions or extras. For two rubrics, a valid example is: {"1":"PRESENT","2":"NOT_PRESENT"}. Verify the key count before responding.

<Prompt>
{query}
</Prompt>

<Response>
{one public final answer}
</Response>

<Rubrics>
1. {rubric description}
2. {rubric description}
...
</Rubrics>
\end{ruactprompt}

For post-training evaluation, we apply the same per-rubric binary judgment instructions and strict JSON output format to frozen model responses, substituting the evaluation query and response for the training inputs. Training-time rubric judgments are provided by Qwen3-32B; benchmark evaluation uses the judges specified in Appendix~\ref{app:evaluation-protocols}.

\end{document}